\pdfoutput=1

\documentclass[11pt]{article}

\usepackage[preprint]{acl}

\usepackage{times}
\usepackage{latexsym}
\usepackage{multirow} 
\usepackage{booktabs}
\usepackage{xcolor}
\usepackage{xspace}
\usepackage{graphicx} 
\usepackage{pifont}
\usepackage{booktabs}
\usepackage{subcaption}
\newcommand{\coloredchange}[2]{%
  {\color{#1}\scriptsize{#2}}%
}

\definecolor{lightred}{RGB}{254,246,245}
\definecolor{lightgreen}{RGB}{240,249,237}
\definecolor{lightblue}{RGB}{238,244,250}
\definecolor{lightcyan}{RGB}{10,110,150}
\hypersetup{
  colorlinks=true,
  linkcolor=lightcyan,
  citecolor=lightcyan,
  filecolor=lightcyan,
  urlcolor=lightcyan
}
\usepackage[capitalize,noabbrev]{cleveref}

\newcommand{\method}{\textsc{NoWait}\xspace}

\usepackage{xcolor}
\usepackage[breakable,skins]{tcolorbox}

\definecolor{boxbackground}{RGB}{255,255,255} 
\definecolor{boxborder}{RGB}{200,200,200}
\definecolor{accentblue}{RGB}{0,114,187}

\newtcolorbox{promptbox}[2][]{ 
    enhanced,
    breakable,
    boxsep=5pt,
    left=9pt,
    right=7pt,
    top=5pt,
    bottom=5pt,
    colback=boxbackground,
    colframe=boxborder,
    boxrule=0.5pt,
    arc=4pt,
    frame hidden, 
    borderline west={3pt}{0pt}{accentblue},
    shadow={0.5pt}{0.5pt}{1.5pt}{black!10},
    fontupper=\normalsize,
    title=#2, 
    colbacktitle=accentblue, 
    coltitle=white,         
    fonttitle={\fontsize{9}{11}\selectfont\bfseries}, 
    attach boxed title to top left={yshift=-2.5mm, xshift=3.2mm},
    boxed title style={
        enhanced,
        left=3pt,
        right=3pt,
        top=1pt,    
        bottom=1pt, 
        boxsep=2pt,
        arc=3pt,
        boxrule=0pt,
        colback=accentblue,
    },
    #1 
}

\newcounter{finding}

\definecolor{lightgray}{RGB}{230,230,230}

\usepackage[T1]{fontenc}

\usepackage[utf8]{inputenc}

\usepackage{microtype}

\usepackage{inconsolata}

\usepackage{graphicx}

\title{\textit{Wait, We Don't Need to ``Wait''}! \\Removing Thinking Tokens Improves Reasoning Efficiency}

\author{
 \textbf{Chenlong Wang},
 \textbf{Yuanning Feng},
 \textbf{Dongping Chen},
 \textbf{Zhaoyang Chu\textsuperscript{1}},
\\
 \textbf{Ranjay Krishna\footnotemark[2]\textsuperscript{2}},
 \textbf{Tianyi Zhou\footnotemark[2]}
\\
\\
 \textsuperscript{1}University College London,
 \textsuperscript{2}University of Washington
}

\begin{document}
\maketitle

{
\renewcommand{\thefootnote}{\fnsymbol{footnote}}
}

\begin{abstract}
Recent advances in large reasoning models have enabled complex, step-by-step reasoning but often introduce significant \textit{overthinking}, resulting in verbose and redundant outputs that hinder efficiency. 
In this study, we examine whether explicit self-reflection, signaled by tokens such as ``\textit{Wait}'' and ``\textit{Hmm}'', is necessary for advanced reasoning.
We propose \method, a simple yet effective approach that disables explicit self-reflection by suppressing these tokens during inference. 
Extensive experiments on ten benchmarks across textual, visual, and video reasoning tasks show that \method reduces chain-of-thought trajectory length by up to 27\%–51\% in five R1-style model series, without compromising model utility. 
\method thus offers a plug-and-play solution for efficient and utility-preserving multimodal reasoning.

\end{abstract}
\section{Introduction}

Recent advancements in large reasoning models (LRMs), exemplified by DeepSeek-R1~\cite{deepseekai2025deepseek-r1}, have shown that complex reasoning abilities can be effectively elicited through simple rule-based reinforcement learning~\cite{qwen3, qwq32b, abdin2025phi4reasoning, coreteam2025mimo}. 
These models produce explicit, step-by-step reasoning through long chain-of-thought (CoT) trajectories~\cite{yang2025early-exit, ma2025nothink} before arriving at final answers.
This capability is believed to be accompanied by the emergence of the ``\textit{Aha Moment}'' phenomenon~\cite{chen2025empirical_r1, yang2025understanding-aha-moment-sexternal}, in which the model begins to rethink problems and self-reflect on its reasoning trajectory with anthropomorphic expressions such as ``\textit{Wait}'', ``\textit{Hmm}'', or ``\textit{Alternatively}''.
This was firstly achieved on R1-style language reasoning models and has been extended to vision-language models (VLMs)~\cite{qvq-72b-preview, kimiteam2025kimi-vl}, enabling multimodal reasoning on images~\cite{zhang2025r1_vl, shen2025vlm_r1, huang2025vision_r1, zhou2025r1_zero_aha_vision} and videos~\cite{feng2025video_r1, qvq-72b-preview, kimiteam2025kimi-vl}.

\begin{figure*}[!t]
	\centering
	\includegraphics[width=\linewidth]{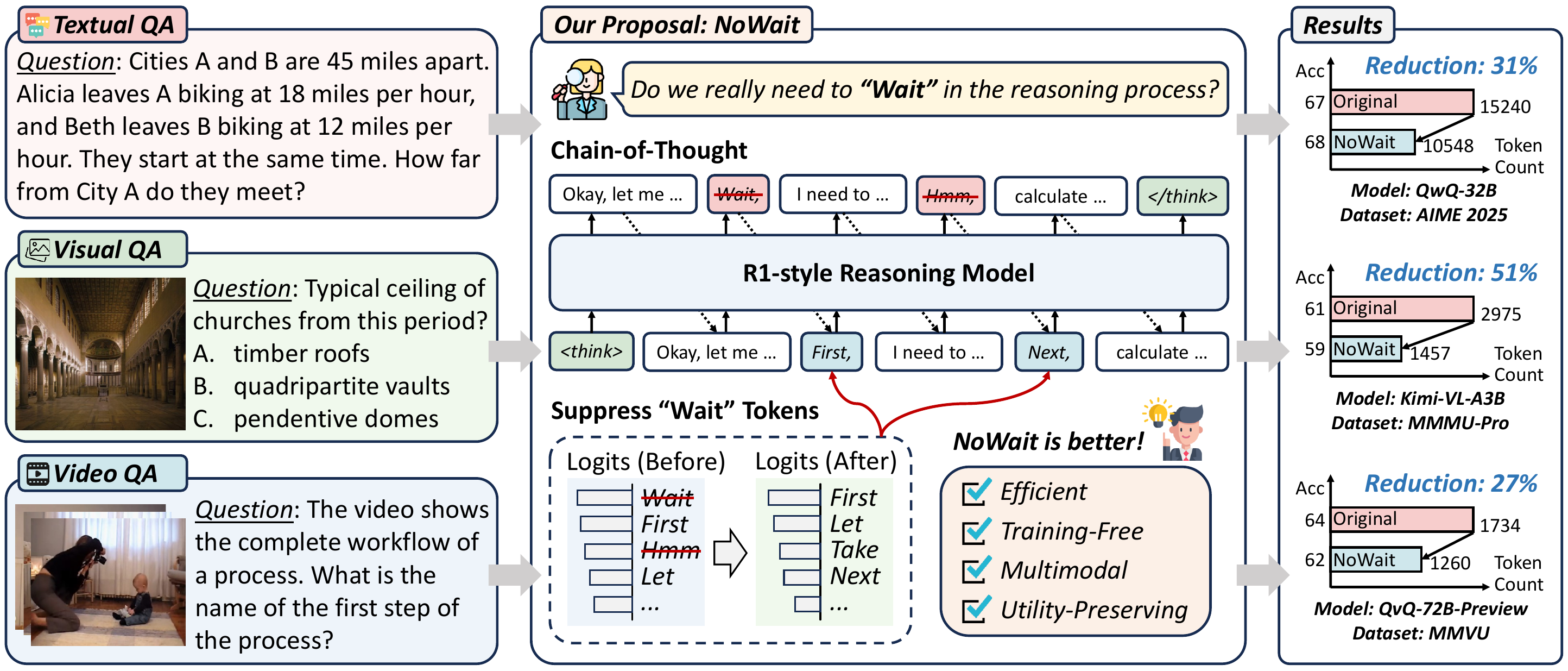}
         \caption{\textbf{Illustrative pipeline for \method.} 
         We introduce \method, a simple yet effective approach that suppresses the generation of reflection keywords (e.g., ``\textit{Wait}'' and ``\textit{Hmm}'') during inference. 
         \method reduces chain-of-thought trajectory length by up to 27\%-51\% across textual, visual, and video reasoning tasks.
         }
	\label{fig_illustration}
\end{figure*}

Despite the effectiveness of long CoT reasoning with self-reflection, the overthinking problem has emerged~\cite{chen2024overthinking, cuadron2025overthinking-danger, chen2025think23overthinking-o1like, wu2025less-understanding-chain-of-thought-length, sui2025stop-overthinking-survey}. 
It is characterized by excessively verbose reasoning and redundant thought steps, often extending over thousands of tokens, resulting in significant computational overhead and high reasoning latency. 
Such inefficiencies hinder the practical deployment of R1-style reasoning models in applications with limited computational resources.

Although numerous efforts have been devoted to efficient reasoning, many existing approaches require additional training, either through reinforcement learning (RL) with length-based rewards~\cite{aggarwal2025l1controllinglongreasoning, liao2025reward-guided-speculative-decoding-efficient, luo2025o1prunerlengthharmonizingfinetuningo1like} or fine-tuning on variable-length CoT trajectories~\cite{ma2025cot_value, munkhbat2025self-training-elicits-concise-reasoning}. 
On the other hand, several training-free approaches have been proposed to mitigate overthinking by reducing token usage during inference. However, they often compromise the overall model utility~\cite{ma2025nothink} or have only demonstrated effectiveness on distilled reasoning models~\cite{yang2025speculative_thinking, yang2025early-exit, xu2025chain-of-draft}.

In this study, we investigate the impact of excessive self-reflection during the reasoning process and question whether explicit self-reflection, signaled by ``\textit{Wait}''-like tokens, is really necessary for advanced reasoning. 
To this end, we propose \method, a simple yet effective training-free approach that disables explicit self-reflection in R1-style reasoning models, significantly reducing token usage while maintaining overall model utility.
As illustrated in~\autoref{fig_illustration}, we directly intervene in the inference process by identifying specific keyword tokens (e.g., ``\textit{Wait}'', ``\textit{Hmm}'', and ``\textit{Alternatively}'') that indicate explicit self-reflection and suppressing their generation.
Specifically, we achieve this by proactively adjusting the logits of these tokens to negative values during decoding, thereby steering the model toward selecting alternative tokens to continue the reasoning process.

Comprehensive experiments show that \method achieves strong performance on ten benchmarks spanning 
\colorbox{lightred}{\ding{182}~\textit{\textbf{textual reasoning}}} (AMC 2023~\cite{ai-mo-amc2023}, AIME 2024, AIME 2025~\cite{maa_aime_problems}, GQPA-D~\cite{rein2024gpqa}), 
\colorbox{lightgreen}{\ding{183}~\textit{\textbf{visual reasoning}}} (MMMU~\cite{yue2023mmmu}, MMMU-Pro~\cite{yue2024mmmu-pro}, MathVista~\cite{lu2024mathvista}, EMMA-mini~\cite{hao2025emma-mini}), and 
\colorbox{lightblue}{\ding{184}~\textit{\textbf{video reasoning}}} (MMVU~\cite{zhao2025mmvu}, VSI-Bench~\cite{yang2024vsi-bench}).
When integrated into five R1-style model series, including QwQ~\cite{qwq32b}, Phi4~\cite{abdin2025phi4reasoning}, Qwen3~\cite{qwen3}, Kimi-VL~\cite{kimiteam2025kimi-vl}, QvQ~\cite{qvq-72b-preview}, \method \textbf{reduces CoT trajectory length by up to 27\%-51\%} across different modalities.
\method serves as a plug-and-play solution for improving reasoning efficiency while preserving overall model utility and provide new insights for the efficient reasoning.

\section{Preliminaries}
\label{preliminary}

\noindent \textbf{Reasoning Model Generation Patterns.}
Reasoning models structure their output using thinking delimiters (i.e., \textit{<think>} and \textit{<\textbackslash think>}), dividing the response into two main components: the CoT trajectory detailing the reasoning process and the final answer summarizing overall thoughts. 

Within the generated CoTs, models employ complex reasoning strategies, such as forward thinking, backtracking, and self-reflection.
Notably, large reasoning models often continue to reason even after obtaining an initial result, performing additional validation steps. Accordingly, we define each segment of reasoning as a \textit{thinking chunk}.
Each thinking chunk is associated with an intermediate answer $r$.
Formally, a thinking chunk can be represented as a pair $(chunk_i, r_i)$, where $chunk_i$ is the reasoning text and $r_i$ is the intermediate answer from $chunk_i$ derived from $chunk_i$. 
Thus, a complete CoT can be structured as follows:
\begin{equation}
    CoT=\{(chunk_i, a_i)\}_{i=1}^n \,.
\end{equation}
The final response is the combination of the CoT trajectory and a concise reasoning summary:
\begin{equation}
    Response=(CoT, summary) \,.
\end{equation}

\noindent\textbf{Self-Reflection within Reasoning Models.}
As stated above, a single CoT can contain multiple reasoning chunks. 
The transitions between these chunks are often marked by specific keywords, such as \textit{Wait}, \textit{Alternatively}, or \textit{Hmm}. 
Models tend to switch their reasoning approaches in subsequent steps, often to verify previous results or explore alternative paths. 
However, this mechanism can sometimes lead to unproductive overthinking, causing models to repeatedly enter new reasoning steps and engage in redundant and unnecessary validation loops.
In this study, we introduce \method, a simple yet effective method for efficient reasoning by intervening in the generation of these keywords.
This method alters models' self-reflection strategies and can be generalized to various modalities.

\section{Removing Thinking Pattern is Better}

In this section, we propose \method, a simple yet effective method, that improves the reasoning efficiency while maintaining acceptable model utility.
We first expand the method details in \autoref{method_nowait}, and introduce the experimental setup in \autoref{method_exp_setup}. We then report the experiment results on \colorbox{lightred}{\textit{\textbf{textual reasoning}}} in \autoref{textual_reasoning}. Additionally, we conduct the comparison experiment in \autoref{comparison-analysis} and further analyze the generalization across different modalities (\colorbox{lightgreen}{\textit{\textbf{visual reasoning}}} and \colorbox{lightblue}{\textit{\textbf{video reasoning}}}) in \autoref{multimodal_reasoning}.

\subsection{Method}
\label{method_nowait}
\method functions as an inference-time intervention.
It \emph{directly prevents} the model from generating the specific tokens associated with self-reflection.
Our method involves three main stages:

\vspace{0.5\baselineskip}
\noindent \textbf{Initialize Reflection Keywords List.} We begin by identifying the initial reflection keywords, such as \textit{``Wait''}, \textit{``Alternatively''}, and \textit{``Hmm''}.
To empirically establish the list, we conduct 32 independent runs of the QwQ-32B~\cite{qwq32b} on AIME 2025~\cite{maa_aime_problems}.
Using ``\textbackslash n\textbackslash n'' as delimiters, we identify the 15 most frequent monolingual words as our identified keywords $K=\{k_i\}$.

\begin{table}[h]
    \centering 
    \label{tab_keywords} 
    \begin{tabular}{lcr}
        \toprule
        \multicolumn{3}{c}{\textbf{Keyword List for Suppressing}} \\
        \midrule
        ``wait'', ``alternatively'', ``hmm'', ``but'', & & \\
        ``however'', ``alternative'', ``another'', & & \\
        ``check'', ``double-check'', ``oh'', & & \\
        ``maybe'', ``verify'', ``other'', ``again'', & & \\
        ``now'', ``ah'', ``any''& & \\
        \bottomrule
    \end{tabular}
\end{table}

\vspace{0.5\baselineskip}
\noindent \textbf{Specific Token-Level Keyword List.} Secondly, for each target model $\alpha$, we expand the initial keyword list $K$ into a specific token-level list, $K_{\alpha}$.
For instance, the variants of ``wait'' include `` wait'', ``Wait'', `` Wait'', ``.wait'' and ``WAIT''.
We achieve this by iterating through the overall vocabulary $V_\alpha$ and identifying all 
variant tokens whose textual representation contains any keyword from $K$ as a substring.
Specifically, we define that, $is\_substr(x, y)=True$ when $x$ is the substring of $y$.
This process can be formulated as follows:
$$K_\alpha=\{v\in V_\alpha | \exists k_s \in K, s.t. is\_substr(k_s, v) \}$$
We further manually filter keywords that are not reasonable (i.e., ``Ohio'' for ``oh'').

\vspace{0.5\baselineskip}
\noindent \textbf{Suppressing Keywords Generation.} During the inference, we leverage a logit processor to prohibit models from generating keywords.
For any keyword $v \in K_\alpha$, its corresponding logit is set to a large negative value.
This effectively makes these reflection-associated tokens, ensuring they are highly unlikely to be sampled by models.

\vspace{0.5\baselineskip}
By surgically preventing the generation of these targeted reflection-associated tokens, \method aims to streamline the LRM's reasoning pathways. This targeted intervention is designed to enhance inference efficiency, reducing both latency and token costs, without requiring any modification to the model's underlying architecture or weights.

\subsection{Experimental Setup}
\label{method_exp_setup}
\noindent \textbf{Model \& Benchmark}.
To comprehensively evaluate the effectiveness of \method, we conduct experiments on the open-source models across different modalities and parameter scales.

For the \colorbox{lightred}{\textit{\textbf{textual reasoning}}} task, we assess reinforcement learning (RL) based models, including QwQ-32B~\cite{qwq32b}, Phi4-Reasoning-Plus~\cite{abdin2025phi4reasoning}, and Qwen3-32B~\cite{qwen3} on math reasoning benchmarks, AIME 2024, AIME 2025~\cite{maa_aime_problems}, and AMC 2023~\cite{ai-mo-amc2023}.

For the \colorbox{lightgreen}{\textit{\textbf{visual reasoning}}} task, our experiments cover the state-of-the-art RL-based visual reasoning models, Kimi-VL-A3B-Thinking~\cite{kimiteam2025kimi-vl} and QvQ-72B-Preview~\cite{qvq-72b-preview} and evaluate on MMMU-Pro~\cite{yue2024mmmu-pro}, MMMU~\cite{yue2023mmmu}, MathVista~\cite{lu2024mathvista} and EMMA-mini~\cite{hao2025emma-mini}

For the \colorbox{lightblue}{\textit{\textbf{video reasoning}}} task, we select QvQ-72B-Preview and evaluate on VSI-Bench~\cite{yang2024vsi-bench} and MMVU~\cite{zhao2025mmvu}.

\begin{table*}[!t]
\setlength{\tabcolsep}{5pt}
\begin{minipage}[t]{\linewidth}
    \caption{\textbf{Experiment results on \colorbox{lightred}{\textit{\textbf{Textual Reasoning}}}Tasks. }}
    \centering
    \begin{tabular}{cllllll}
        \toprule
        \multirow{2}{*}{\textbf{Strategy}} & \multicolumn{2}{c}{\textbf{AMC 2023}} & \multicolumn{2}{c}{\textbf{AIME 2024}} & \multicolumn{2}{c}{\textbf{AIME 2025}} \\
         & ACC$\uparrow$ & LEN$\downarrow$ & ACC$\uparrow$ & LEN$\downarrow$ & ACC$\uparrow$ & LEN$\downarrow$ \\
         \midrule
         \midrule
         \multicolumn{7}{c}{\textbf{QwQ-32B}} \\
         \midrule
         \textbf{Original} & 91.25 & 7542 & 73.33 & 14142 & 66.67 & 15240 \\
         \textbf{NoThink} & 72.50 & 4265 & 46.67 & 7980 & 40.00 & 8167 \\
         \textbf{\method} & 
         95.50  \coloredchange{red}{+4.25} & 
         5267   \coloredchange{blue}{-30\%} & 
         71.33      \coloredchange{red}{-2.00} & 
         11907      \coloredchange{blue}{-16\%} & 
         68.00  \coloredchange{red}{+1.33} & 
         10548  \coloredchange{blue}{-31\%} \\
         
         \midrule
         \multicolumn{7}{c}{\textbf{Phi4-Resoning-Plus}} \\
         \midrule
         \textbf{Original} & 90.00 & 6366 & 70.00 & 15161 & 59.33 & 16257 \\
         \textbf{NoThink} & 80.83 & 3805 & 34.67 & 6200 & 31.33 & 5549 \\
         \textbf{\method} & 
         96.00  \coloredchange{red}{+6.00} & 
         4524   \coloredchange{blue}{-28\%} & 
         69.33      \coloredchange{red}{-0.67} & 
         11185      \coloredchange{blue}{-26\%} &
         62.67  \coloredchange{red}{+3.34} &
         12490  \coloredchange{blue}{-23\%} \\

         \midrule
         \multicolumn{7}{c}{\textbf{Qwen3-32B}} \\
         \midrule
         \textbf{Original} & 97.50 & 6424 & 81.33 & 12720 & 66.67 & 14987 \\
         \textbf{NoThink} & 59.50 & 1240 & 25.33 & 2511 & 20.00 & 2165 \\
         \textbf{\method} & 
         96.67  \coloredchange{red}{-0.83} & 
         5560   \coloredchange{blue}{-13\%} & 
         83.33  \coloredchange{red}{+2.00} & 
         10732  \coloredchange{blue}{-16\%} &
         64.44  \coloredchange{red}{-2.67} &
         12930  \coloredchange{blue}{-14\%} \\

        \bottomrule

    \end{tabular}
\label{tab_text_reasoning}
\end{minipage}
\end{table*}

\vspace{0.5\baselineskip}
\noindent \textbf{Metrics.}
The goal of \method is to preserve the model's reasoning accuracy while substantially diminishing the number of generated tokens during inference. 
Performance is assessed using two key metrics: 
\ding{182} \textbf{\textit{Accuracy} (ACC)}: This measures the correctness of the model's final output.
\ding{183} \textbf{\textit{Generation Length} (LEN)} quantifies the average number of tokens generated by the model per problem instance, calculated over $n$ independent runs.

\vspace{0.5\baselineskip}
\noindent \textbf{Baselines.} To compare the latency of \method, we use both the models' original performance and the NoThink strategy~\cite{ma2025nothink} as baselines. 
Both \method and NoThink share a similar rationale, aiming to intervene in the model's reasoning process. 
However, while \method operates at the token level by prohibiting the output of specific tokens, NoThink attempts to directly remove the entire reasoning process by prompt engineering.
By including these baselines, we can conduct a more comprehensive analysis about the model's performance under different intervention approaches.

\vspace{0.5\baselineskip}
\noindent \textbf{Experiment Details.}
For each evaluated benchmark, we conduct five independent runs.
Except for the Qwen3 series, we infer without chat templates on open-ended problems and leverage the same prompt template for multiple-choice problems (see \autoref{multiple_choice_question_prompt_template}).
Because of the different thinking patterns, we apply chat templates for the Qwen3 model inference.
In baseline and \method experiments, we set a maximum token limit of 32,768 tokens per instance.
If a model's generation reaches this limit before finishing CoT generation, that instance is considered incorrect, and the generation length is 32,768 tokens.
If not, we will extract the final answer from the generated CoT and judge the correctness.
This policy ensures that models failing to complete their response within the budget are appropriately penalized in \textit{Accuracy} metric.
For NoThink strategy~\cite{ma2025nothink}, we set a token budget of 10,000. 
Details of the token budget applied for NoThink can be found in \autoref{nothink_implementation}.

\subsection{LRMs can be Efficient without ``\textsc{Wait}''}
\label{textual_reasoning}
\autoref{tab_text_reasoning} presents a comprehensive quantitative overview of our \method's performance on various \colorbox{lightred}{\textit{\textbf{textual reasoning}}} tasks, evaluated across different LRMs with diverse model structures and parameter scales.
Our method \method consistently and significantly reduces the output length while maintaining the reasoning accuracy.

\vspace{0.5\baselineskip}
\noindent \textbf{Model Architectures Generalization}.
\label{architecture_generalization}
Notably, when integrated with QwQ-32B, \method improves accuracy on AMC 2023 by 4.25 percentage points, while reducing output length to just 70\% of the baseline.
With another model architecture, Phi4-Reasoning-Plus, our method achieves an even greater improvement of 6.00 percentage points, alongside a 28\% reduction in token generation.
Additionally, Qwen3-32B also benefits from our approach, reducing output length by 13\% with only a marginal decrease in reasoning accuracy.
These results demonstrate that our method \method consistently enhances efficiency across diverse model architectures.
This consistency suggests a fundamental similarity in the reasoning patterns and redundancy present in different models, underscoring the broad applicability of our approach.

\begin{table}[!t]
\setlength{\tabcolsep}{4pt}
    \caption{\textbf{Comparison Experiments across Multiple Efficient Reasoning Methods.} We use QwQ-32B-Preview for experiments.}
    \centering
    \begin{tabular}{l|llll}
        \toprule
        \multirow{2}{*}{\textbf{Strategy}} & \multicolumn{2}{c}{\textbf{AIME 2024}} & \multicolumn{2}{c}{\textbf{AMC 2023}}
        \\
         & ACC$\uparrow$ & LEN$\downarrow$ & ACC$\uparrow$ & LEN$\downarrow$ \\
         \midrule
         \midrule
         Baseline 
         & 42.00
         & 8979
         & 82.50
         & 4143 
         \\
         \midrule
         Token-Budget 
         & 46.67
         & 8734
         & 82.50
         & 3636 
         \\
         \midrule
         O1-Pruner 
         & 33.33
         & 4289
         & 77.50
         & 2399
         \\
         \midrule
         \method 
         & 42.00
         & 5764
         & 86.00
         & 3396 
         \\
         \bottomrule
    \end{tabular}
    \label{tab_comparison}
\end{table}

\vspace{0.5\baselineskip}
\noindent \textbf{Reasoning Difficulty Analysis}.
\label{difficulty_generalization}
We tested our method on mathematical reasoning benchmarks spanning various difficulty levels (AMC 2023 $<$ AIME 2024 $<$ AIME 2025).
The experimental statistics demonstrated strong generalization across these levels:
All tested models achieved comparable reductions in token usage regardless of task difficulty.
Crucially, \method enabled models to maintain or even improve performance on more challenging tasks.
For instance, QwQ-32B achieved a 1.33\% point increase on the challenging AIME 2025 benchmark, while reducing token usage by 31\%, which is comparable to its performance on the college-level AMC 2023.
Qwen3-32B consistently reduced output length by 14\% to 16\% across all three math benchmarks, while Phi4-Reasoning-Plus showed similar gains and reductions from 23\% to 28\%.
On the non-mathematical GPQA-Diamond task, models showed a slight performance decrease compared to the math reasoning benchmarks, but still maintained efficiency, with an overall 11.67\% reduction in token usage.

\vspace{0.5\baselineskip}
These consistent efficiency gains and stable performance across diverse models and varying tasks suggest that, despite the architecture and scale, LRMs exhibit similar inherent redundancy in their reasoning processes. \method effectively prunes this redundancy, demonstrating that substantial efficiency improvements can be achieved simply by suppressing the keywords generation, without the need for complex explicit ``waiting'' mechanisms.

\subsection{Comparison Analysis}
\label{comparison-analysis}
\noindent \textbf{Comparison Experiment.}
We further compare with existing efficient reasoning techniques, including prompt-based training-free technique, Token-Budget~\cite{han2024token-budget}, and training-based technique, O1-Pruner~\cite{luo2025o1prunerlengthharmonizingfinetuningo1like}, using QwQ-32B-Preview~\cite{qwq32b} on AIME 2024 and AMC 2023.
All inference is conducted without chat templates to ensure fairness.

\begin{table*}[!t]
\setlength{\tabcolsep}{4pt}
\begin{minipage}[t]{\linewidth}
    \caption{\textbf{Experiment results on \colorbox{lightgreen}{\textit{\textbf{Visual Reasoning}}} Tasks.}}
    \centering
    \begin{tabular}{cllllllll}
        \toprule
        \multirow{2}{*}{\textbf{Strategy}} & \multicolumn{2}{c}{\textbf{MMMU-Pro}} & \multicolumn{2}{c}{\textbf{MMMU}} & \multicolumn{2}{c}{\textbf{MathVista}} & \multicolumn{2}{c}{\textbf{EMMA-mini}}\\
         & ACC$\uparrow$ & LEN$\downarrow$ & ACC$\uparrow$ & LEN$\downarrow$ & ACC$\uparrow$ & LEN$\downarrow$ & ACC$\uparrow$ & LEN$\downarrow$ \\
         \midrule
         \midrule
         \multicolumn{9}{c}{\textbf{Kimi-VL-A3B-Thinking}} \\
         \midrule
         \textbf{Original} & 61.27 & 2975 & 57.00 & 2929 & 71.50 & 1822 & 34.75 & 5734 \\
         \textbf{\method}
         & 58.73 \coloredchange{red}{-2.54}
         & 1457 \coloredchange{blue}{-51\%}
         & 55.20 \coloredchange{red}{-1.80}
         & 1746 \coloredchange{blue}{-40\%}
         & 69.40 \coloredchange{red}{-2.10}
         & 1045 \coloredchange{blue}{-43\%}
         & 27.50 \coloredchange{red}{-7.25}
         & 2269 \coloredchange{blue}{-60\%}
         \\

         \midrule
         \multicolumn{9}{c}{\textbf{QvQ-72B-Preview}} \\
         \midrule
         \textbf{Original} & 65.77 & 2094 & 66.85 & 1977 & 73.54 & 1338 & 32.00 & 2097 \\
         \textbf{\method}
         & 63.79 \coloredchange{red}{-1.98}
         & 1659  \coloredchange{blue}{-21\%}
         & 66.74 \coloredchange{red}{-0.11}
         & 1571  \coloredchange{blue}{-21\%}
         & 70.92 \coloredchange{red}{-2.62}
         & 939   \coloredchange{blue}{-30\%}
         & 28.00 \coloredchange{red}{-4.00} 
         & 1554  \coloredchange{blue}{-26\%}
         \\
         \bottomrule
    \end{tabular}
\label{tab_multi_modality}
\end{minipage}
\end{table*}

\begin{table}[!t]
\label{tab_video_result}
\setlength{\tabcolsep}{4pt}
    \caption{\textbf{Experiment Results on \colorbox{lightblue}{\textit{\textbf{Video Reasoning}}} Tasks.} We use QvQ-72B-Preview for experiments.}
    \begin{tabular}{c|llll}
        \toprule
        \multirow{2}{*}{\textbf{Strategy}} 
        & \multicolumn{2}{c}{\textbf{MMVU}} 
        & \multicolumn{2}{c}{\textbf{VSI-Bench}} \\
        & ACC$\uparrow$ & LEN$\downarrow$ & ACC$\uparrow$ & LEN$\downarrow$\\
        \midrule
        \midrule
        \textbf{Original} & 64.10 & 1734 & 22.51 & 1280\\
        \textbf{\method}
        & 62.20    
        & 1260     
        & 22.57    
        & 1020     
        \\
        \midrule
        \textbf{Performance} & -1.90 & -27\% & +0.06 & -20\% \\
        \bottomrule
    \end{tabular}
\label{tab_video_result}
\end{table}

As shown in \autoref{tab_comparison}, \method exhibits more significant generation length curtailment compared to Token-Budget.
Although Token-Budget shows promising results on base models, such as GPT-4o, its effectiveness does not generalize to current LRMs(Deepseek-R1~\cite{deepseekai2025deepseek-r1}, QwQ-32B~\cite{qwq32b}).
These reasoning models are less sensitive to the prompt design, resulting in less efficiency.
O1-Pruner, while effective at reducing token usage, incurs severe performance degradation on QwQ-32B-Preview.
In contrast, \method does not require additional training or data, but instead guides models to strike an effective balance between output length and reasoning accuracy, achieving a spontaneous trade-off.

\vspace{0.5\baselineskip}
\noindent \textbf{LRM Cannot Skip Thinking}.
\label{nothink_anaylsis}
As shown in \autoref{tab_text_reasoning}, Qwen3-32B, a model specifically trained for non-thinking patterns, exhibits notable reductions in token usage.
However, for other models (QwQ-32B and Phi4-Reasoning-Plus) without non-thinking pattern training, NoThink~\cite{ma2025nothink}, a prompt-based method, fails to thoroughly skip the generation of reasoning steps.
While NoThink does reduce the generation length, the evaluated model can still generate the thinking process and demonstrate a serious compromise in accuracy.
This failure indicates that the presence of explicit "thinking" tokens (``<think>'' and ``<\textbackslash think>'') can influence models' output, but is insufficient to precisely control models' reasoning strategy.
Our method \method operates on a similar premise by targeting key reasoning-related tokens, but achieves significant efficiency improvements with better maintain on reasoning accuracy.

\subsection{Efficient Multimodal Reasoning}
\label{multimodal_reasoning}
In this section, we propose efficient multimodal reasoning.
We assess \method on visual reasoning models using image and video reasoning benchmarks.
As shown in \autoref{tab_multi_modality} and \autoref{tab_video_result}, visual reasoning models exhibit more exciting outcomes.

\vspace{0.5\baselineskip}
\noindent \textbf{Severe Verbosity on Multimodal Reasoning}.
Although Kimi-VL-A3B-Thinking generates an average of only 2,000 tokens across four image reasoning benchmarks, significantly fewer than that in math reasoning tasks, our method \method further reduces the generation length by an average of 49\%, with only a modest overall accuracy drop of 3.42 percentage points.
A similar trend is observed with QvQ-72B-Preview, which achieves up to a 30\% reduction in token usage, accompanied by only a slight decrease in accuracy (ranging from 0.11\% to 4.00\%).
For video reasoning tasks, QvQ-72B-Preview also demonstrates substantial reductions in output length while maintaining comparable accuracy. 
Similar to textual reasoning tasks, these results reveal the same challenging problems that a significant portion of generated tokens are either redundant or contribute little to the final reasoning.
Existing multimodal reasoning models still suffer from severe reasoning inefficiency.

\vspace{0.5\baselineskip}
\noindent \textbf{Reinforcement Learning is Less Efficient.}
\label{rl_is_less_efficient}
We further evaluate various RL-based reasoning models across varying benchmarks and modalities.
While a generation of intellectual reasoning models confirms the effectiveness of the RL algorithm in advanced reasoning capabilities, the efficiency of the optimal policy derived from the RL algorithm is still disappointing.
The model learns a reasoning policy from training and begins to spontaneously reflect reasoning processes during inference.
However, these algorithms fail to effectively teach models when reflection is truly necessary.
As a result, these models often adopt a lower threshold for self-reflection, leading to unnecessary verification steps and less efficient reasoning.
Our method suppresses the generation of reflection keywords, raising the threshold of self-reflection, and making it more efficient and necessary.

\begin{figure*}[!t]
    \centering
    \includegraphics[width=0.99\textwidth]{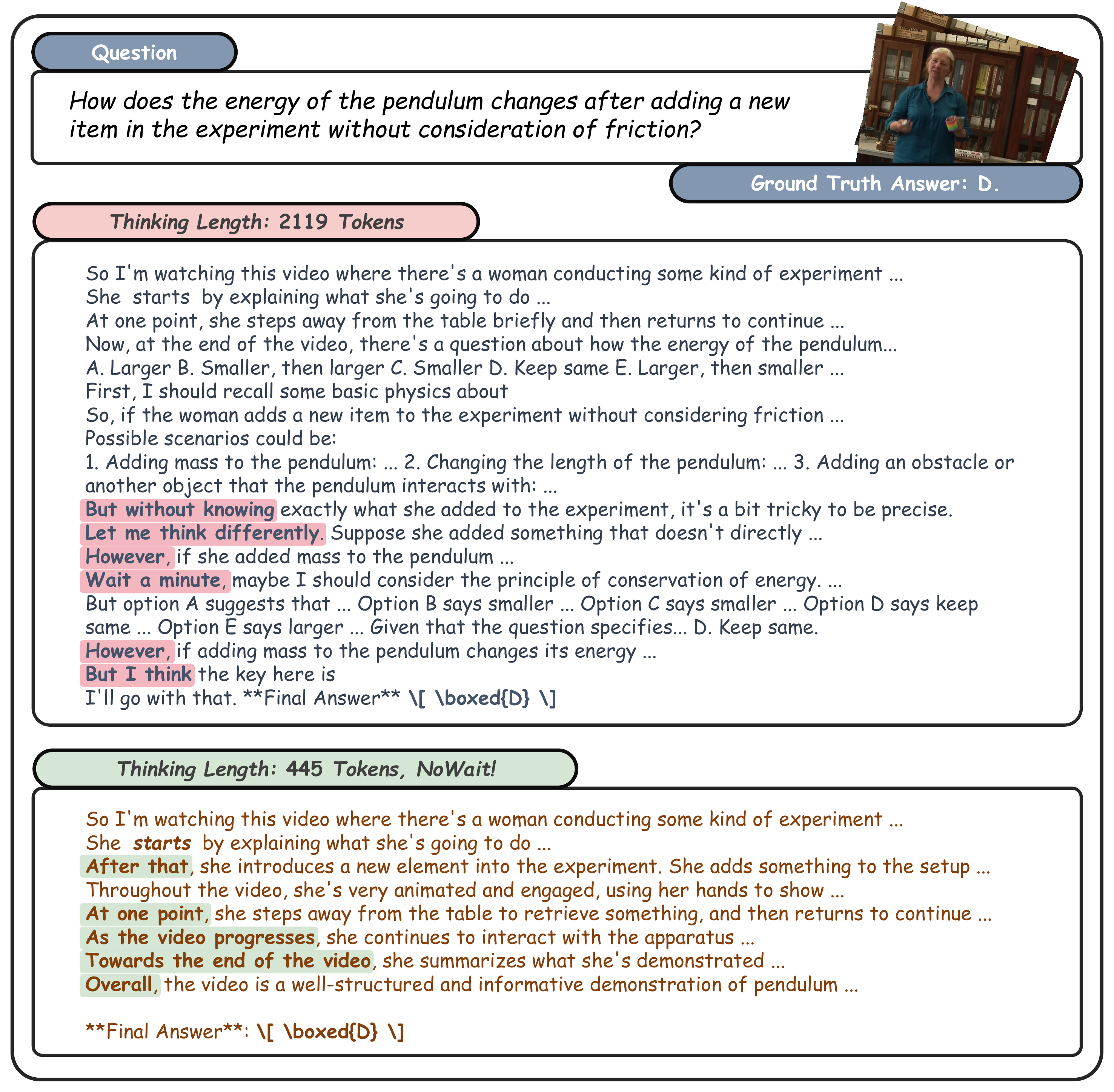}
    \caption{\textbf{One Case Study From QvQ-72B-Preview on MMVU.} \method CoT is more straightforward than the original CoT, without unnecessary self-reflection and verbosity.}
    \label{fig:video_NoWait_Cot}
\end{figure*}

\section{Discussion}

\begin{figure}[!t]
    \centering
    \includegraphics[width=\linewidth]{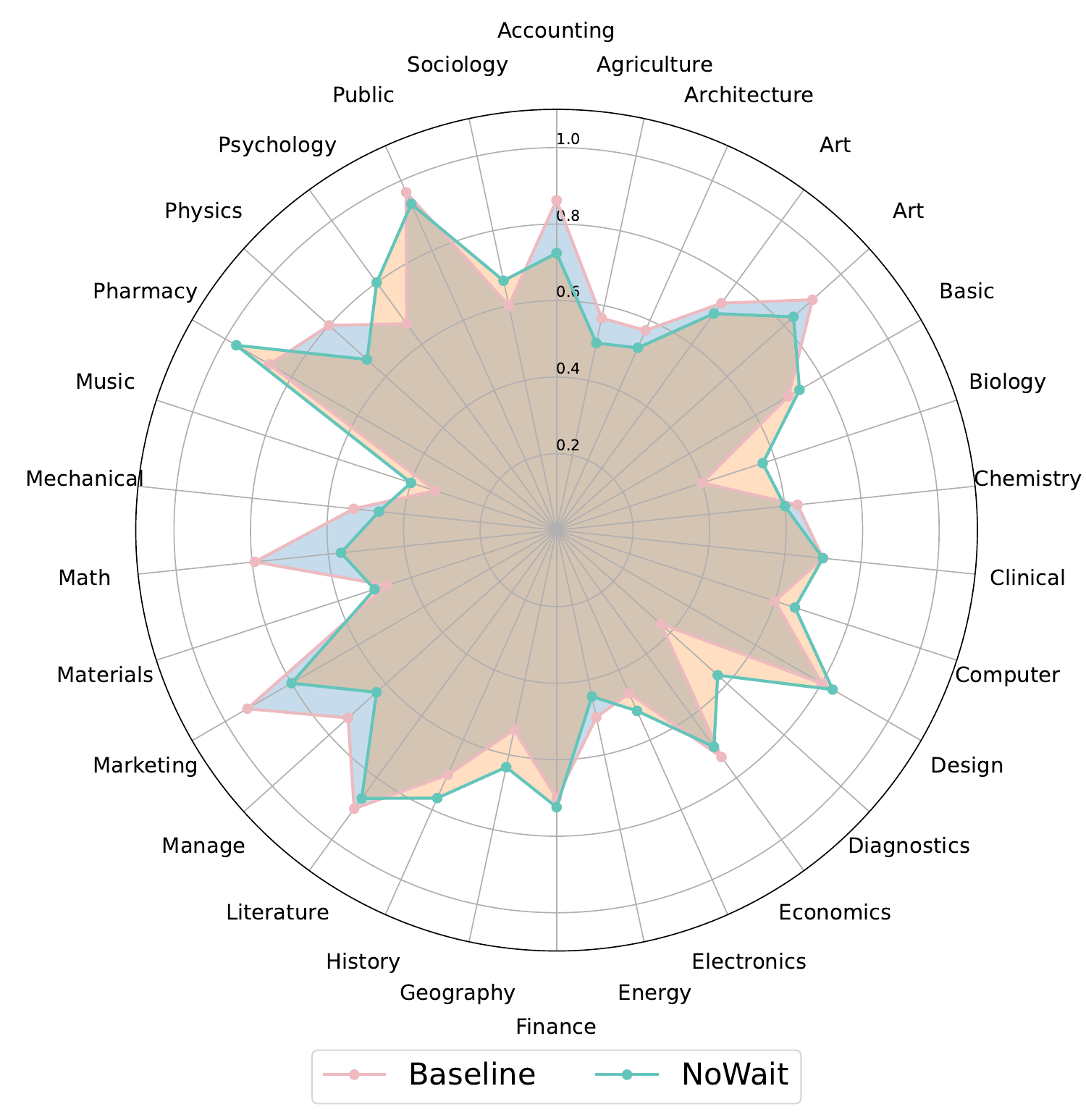} 
    \caption{\textbf{Accuracy Radar Map on MMMU for QvQ-72B-Preview.}}
    \label{fig_mmmu_rader}
\end{figure}
\vspace{-0.5\baselineskip}

In this section, we first discuss the effectiveness of our method \method in \autoref{why_work} by case study and the robustness of the model while applying \method in \autoref{discussion_robustness}.
Additionally, we conduct an empirical experiment to analyze the difference between RL-based models and distill models based on \method in~\autoref{discussion_distill}.

\subsection{Why does \method Work?}
\label{why_work}
As we discussed in \autoref{nothink_anaylsis}, thinking tokens (``<think>'' and ``<\textbackslash think>'') cannot thoroughly control models' actions.
Similarly, can banning keywords completely remove self-reflection from CoTs?
If not, why \method result in more efficient reasoning?
To answer this, we conduct a case study to analyze the effectiveness of our method.

\vspace{0.5\baselineskip}
\noindent \textbf{More Efficient Self-Reflection Mechanism}.
\label{case_study_2_effecient_reflection}
\method does not prohibit models from self-reflection.
However, this method guides models to skip the unnecessary ``waiting'' reasoning.
To illustrate this, we select an example from Qwen3-32B on AMC 2023 benchmark.
The \method CoT (see \autoref{fig:AMC23-NOWAIT1}) is noticeably shorter than the original CoT (see \autoref{fig:AMC23-original1}).
Specifically, the \method CoT reserves its self-reflection for two clear points: \ding{182} it notes the extraneous root and instantly discards it after factoring, \ding{183} it quickly verifies both original equations with the numeric solution.
By contrast, the original CoT continually interjects ``let me check again'', and ``perhaps another way'', leading to 5 derivations of essentially the same algebra.
In short, the first approach builds a more concise reasoning process with necessary checks to ensure correctness, whereas the second strategy prefers to pause to highlight every minor thought, making the logic scattered and less efficient.

\vspace{0.5\baselineskip}
\noindent \textbf{Concise and Straightforward Reasoning}.
\label{case_study_1_concise}
\autoref{fig:video_NoWait_Cot} presents an example from QvQ-72B-Preview on MMVU.
The original CoT contains six instances of self-reflection, resulting in excessive token usage and a disorganized reasoning process.
In contrast, the \method CoT exhibits a more streamlined and coherent approach.
The model analyzes the video in detail, using a series of time-sequence cues such as ``starts'', ``After that'', ``At one point'', ``As the video progresses'' and ``Towards the end of the video''.
With fewer self-reflections, the \method CoT organizes its reasoning more logically and systematically, whereas the original CoT appears fragmented and less focused, always generating a new reasoning branch by ``Wait''.
Ultimately, the \method derives the final answer directly from its detailed analysis.
Unlike the original policy, \method encourages the model to connect observations to conclusions more directly, reducing unnecessary speculation and making the reasoning process more concise and straightforward.

\subsection{A Closer Look at RL Models Performance}
\label{discussion_robustness}
For textual reasoning tasks, our evaluation primarily focuses on the math problems.
As we discussed in \autoref{difficulty_generalization}, \method yields consistent experimental outcomes across math benchmarks of varying difficulty levels.
For multimodal reasoning tasks, \autoref{fig_mmmu_rader} shows the accuracy of the QvQ-72B-Preview on MMMU across a wide range of fields.
A crucial observation highlights remarkably small accuracy divergence between the baseline and \method in almost all tested disciplines. 
Despite the potential intervention introduced by \method, the model's performance remains closely aligned with the baseline across diverse academic and professional subjects. 
This minimal degradation strongly indicates the robustness of the QvQ-72B-Preview when applying \method, highlighting generalization capability across varying areas.

\subsection{Distilled Models Cannot Reasoning without ``Wait''}
\label{discussion_distill}
Recent studies~\cite{yue2025doesreinforcementlearningreally} underscore the significant differences between reasoning models based on reinforcement learning (RL) and those trained through distillation.
To better understand these differences, we further evaluate the effectiveness of \method across Qwen3 series, including an RL-based model (Qwen3-32B) and several distilled models (Qwen3-4B/8B/14B).

\begin{figure}[!t]
    \centering
    \includegraphics[width=\linewidth]{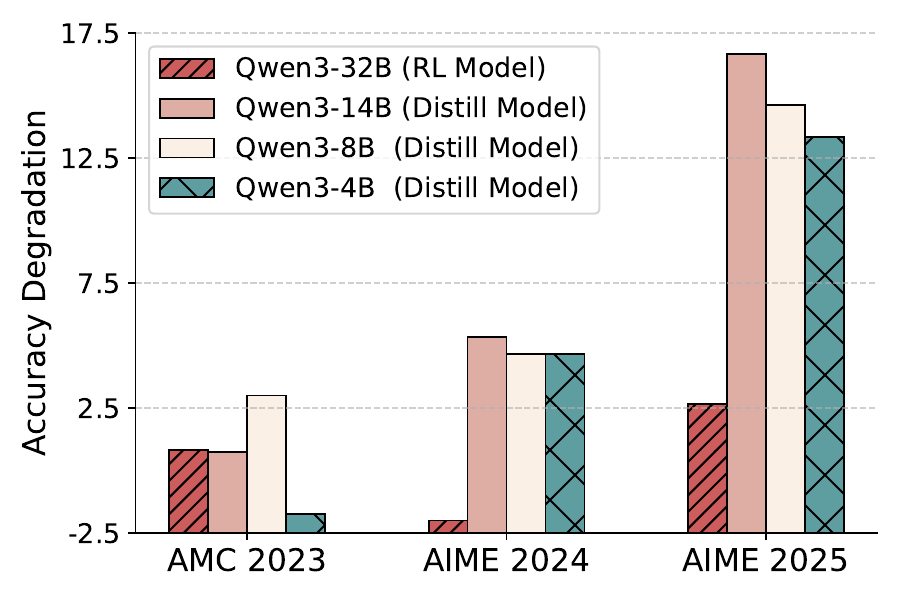}
    \caption{\textbf{Accuracy Degradation across Qwen3 Seires Models on Math Reasoning Benchmarks.}}
    \label{fig_distill}
\end{figure}

\autoref{fig_distill} illustrates the accuracy degradation for models using \method, where a higher score indicates a more pronounced decline.
The selected math reasoning benchmarks differ in difficulty, ordered as follows:
AMC 2023 < AIME 2024 < AIME 2025.
While the RL-based models maintain consistent performance across these benchmarks, distilled models exhibit a distinct trend of increasing accuracy degradation as difficulty rises.

Specifically, distilled models show similar accuracy degradation relative to the RL-based model on the simpler AMC 2023.
This performance gap extends significantly as problem difficulty increases, surpassing a 5-percentage-point drop on AIME 2024 and dramatically exceeding 12 percentage points on more challenging AIME 2025.

This sharp performance degradation among distilled models, in contrast to the stable performance of the RL-based model, demonstrates their higher sensitivity to reflection keywords.
Given that the supervised fine-tune (SFT) directly injects new knowledge into models, the CoT structures are crucial for advanced reasoning.
Simply removing these keywords, however, severely disrupts the inherent CoT structure, restricting distilled models from exhibiting full reasoning capabilities. 
Especially on more challenging reasoning problems, distill models fail to effectively conduct validation, suffering from substantial underthinking.

\section{Conclusion}

This work demonstrates that explicit self-reflection, signaled by tokens such as ``\textit{Wait}'' and ``\textit{Hmm}'', is not essential for advanced reasoning in R1-style models. 
By suppressing these tokens during inference, the proposed \method approach effectively reduces overthinking and shortens chain-of-thought trajectories without compromising overall model utility. 
Extensive experiments across diverse models and benchmarks in textual, visual, and video reasoning tasks demonstrate that \method serves as an efficient and utility-preserving solution for multimodal reasoning,  offering new insights for the lightweight deployment of large reasoning models.

\section*{Acknowledgment}
Many thanks to Yao Wan and Jieyu Zhang for their invaluable support and comments.

\section*{Limitation}
In this study, we introduce \textsc{NoWait}, a simple yet effective method for efficient reasoning on all modalities.
We conduct experiments across a large range of models across various benchmarks to validate the effectiveness of our method.
Although the promising results, we acknowledge that existing benchmarks cannot comprehensively exhibit the reasoning capabilities of models from all aspects.

\bibliography{custom}

\appendix
\section{Related Work}
\paragraph{Large Reasoning Model} 
The pursuit of advanced reasoning capabilities in Large Language Models (LLMs)~\cite{openai2024gpt4ocard} has spurred significant research, particularly focusing on strategies that scale computation~\cite{chen2024test-time-scaling, snell2024test-time-scaling} or refine the generation process during inference.
Prior studies apply fundamental techniques like Chain-of-Thought (CoT) prompting~\cite{wei2023chain-of-thought-key}, guiding the model to think step by step, or integrate Process Reward Models(PRMs), external verifiers, and search-guided decoding~\cite{brown2024largelanguage-monkeys-scaling} to aggregate multiple reasoning paths and enhance final answer accuracy.
These efforts have culminated in a new generation of powerful Large Reasoning Models (LRMs), such as ChatGPT-O1~\cite{chatgpt-o1}, Deepseek-R1~\cite{deepseekai2025deepseek-r1}, QwQ~\cite{qwq32b}, Gemini2.5~\cite{gemini2.5pro}, which enable to spontaneous generation of extensive CoT sequences involving forward thinking, backtracking, and verification steps.
Within the open-source domain, models derive reasoning abilities from diverse training paradigms, primarily through reinforcement learning (RL)~\cite{deepseekai2025deepseek-r1, ramesh2024grpo, muennighoff2025s1-simple-test-time-scaling} on reasoning tasks or distillation~\cite{deepseekai2025deepseek-r1, yu2024distilling21} on high-quality CoT data produced from RL-based models.
Recent works~\cite{yue2025doesreinforcementlearningreally} have analyzed the difference between the two types of models.
In this study, we include RL-based models for further exploration, underscoring the defects of RL-triggered reasoning capabilities.

\paragraph{Efficient Reasoning}
While elaborating reasoning processes like long CoT demonstrates enhanced performance on reasoning tasks, the associated verbosity presents a significant efficiency challenge~\cite{chen2024overthinking}.
The generation of extensive intermediate steps substantially increase inference latency and computational cost, hindering practical deployment in real-world applications.
Consequently, a considerable body of work explores methods for efficient reasoning, aiming to reduce the length of reasoning traces without compromising accuracy.
Some techniques continue to train models for CoT optimization~\cite{aggarwal2025l1controllinglongreasoning, luo2025o1prunerlengthharmonizingfinetuningo1like, shen2025dast-difficultyadaptiveslowthinkinglarge}, such as applying RL with length-based reward design~\cite{sun2024fast-best-of-n-decoding-speculative, liao2025reward-guided-speculative-decoding-efficient, luo2025o1prunerlengthharmonizingfinetuningo1like, aggarwal2025l1controllinglongreasoning}, or fine-tuning with variable-length CoT data~\cite{han2024token-budget, yu2024distilling21, munkhbat2025self-training-elicits-concise-reasoning}.
Other methods conform training-free strategy, applying dynamic reasoning paradigms during inference~\cite{yang2025early-exit, zhang2025probing-hidden-state, wu2025less-understanding-chain-of-thought-length, lin2025sleep-time-compute-inference-scaling} or leveraging prompts to guide efficient reasoning~\cite{cheng2024ccot, xu2025chain-of-draft, han2024token-budget, ma2025nothink}.
While existing studies are effective in cutting down the token usage, our study provides a new insight to rethink the internal mechanism of efficient reasoning and propose efficient multimodal reasoning.

\paragraph{Self-Reflection \& Overthinking}
Parallel to enhancing reasoning capabilities and efficiency, recent studies analyze the intricacies of the generated thought processes.
Within these generated sequences, an interesting phenomenon occurs - moments marked by keywords like "wait" and "hmm", which we term \textit{Aha Moment}~\cite{deepseekai2025deepseek-r1, liu2025understanding-r1-zero-like-training-critical}.
These moments seemingly indicate a capability for self-reflection~\cite{chen2025reasoning-era-survey-long}, allowing models to reassess their reasoning path and verify their CoT before concluding.
Prior studies~\cite{yang2025understanding-aha-moment-sexternal, zhang2025probing-hidden-state} have begun to characterize these moments and probe the latent states to explore the potential mechanisms behind such spontaneous self-reflection.
However, the frequent occurrence of these keywords can also lead to significant \textit{Overthinking}~\cite{chen2024overthinking, sui2025stop-overthinking-survey}, where the model continues reflecting even after reaching correct intermediate or final conclusions. 
Building on the initial characterizations from previous work, our study takes a further step to evaluate the functional effectiveness of these spontaneously generated Aha Moments, directly addressing whether they are essential contributors to the reasoning outcomes or potentially represent a form of inefficient behavioral mimicry.

\section{Baseline Implementation Details}
\label{baseline-implementation}
Our experiments include three baselines, NoThinking~\cite{ma2025nothink}, TokenBudget~\cite{han2024token-budget}, and O1-Pruner~\cite{luo2025o1prunerlengthharmonizingfinetuningo1like}.
In this section, we will systematically introduce the implementation details of these techniques.

\subsection{NoThinking}
\label{nothink_implementation}
The core idea of NoThinking is to leverage prompts, guide reasoning models to skip the reasoning processes, and directly generate a final response.
For models that have not been post-trained for non-reasoning mode, such as QwQ-32B and Phi4-Reasoning-Plus, we apply the prompt template as follows:

\begin{tcolorbox}[colback=gray!20, colframe=gray!50!black, title=Prompt Template for NoThinking]
\{Question\}

<think>

Okay, I think I have finished thinking.

<\textbackslash think>
\end{tcolorbox}

We then adopt a budget forcing technique specifically for NoThinking.
Different from the token budget we apply for normal inference and \method,  we set the token budget to 10,000 and forced models to generate \textit{Final Answer} when the model reaches the token budget.

\subsection{Token-Budget}
\label{token-budget_implementation}
We apply the TALE-EP strategy, a prompt-based method.
This method consists of two steps:

\ding{182} Directly answering the reasoning model: 
\begin{tcolorbox}[colback=gray!20, colframe=gray!50!black, title=Prompt Template for TALE-EP]
Task: Analyze the given question and estimate the minimum number of tokens required to generate a complete and accurate response. Please give the
response by strictly following this format: [[budget]],
for example, Budget: [[12]].
\end{tcolorbox}

\ding{183} We include a token budget in the prompt to guide models to think efficiently.
\begin{table}[htp]
    \centering
    \caption{Prompt Template Applied for Token Budget.}
    \begin{tabular}{c|l}
         \toprule
         \textbf{Prompt method} & \textbf{Content} \\
         \midrule
         Vanilla CoT & Let's think step by step: \\
         \midrule
         \multirow{2}{*}{Token Budget} & Let's think step by step and\\
         & use less than \{budget\} tokens:\\
         \bottomrule
    \end{tabular}
    \label{tab:my_label}
\end{table}

\subsection{O1-Pruner}
\label{o1-pruner-implementation}
O1-Pruner is an effective post-training method.
We select a released model trained on QwQ-32B-Preview by O1-Pruner.
This model can be accessed via \href{https://hf-mirror.com/LordNoah/QwQ-32B-Preview-Pruned}{Hugging Face}.

\section{Prompts}
\label{multiple_choice_question_prompt_template}
\begin{tcolorbox}[colback=gray!20, colframe=gray!50!black, title=Prompt Template for \\ Multiple-Choice Question]
\{Question\}

---

Choices:\\
A. option A \\
B. option B \\
... \\
--- \\
Choose the correct answer from the choices above. \\
Output format: [ANSWER: "<answer>"] If the answer is A, output [ANSWER: "A"] \\
\end{tcolorbox}

\section{Benchmark \& Models}

\subsection{Textual QA}
In this paper, we evaluate a range of mathematics competition benchmarks designed to assess the mathematical reasoning abilities of models, including \textbf{AIME2024}, \textbf{AIME2025}, \textbf{AMC2023}. We have also evaluated \textbf{GPQA-Diamond}~\cite{rein2024gpqa}, a challenging benchmark spanning biology, physics, and chemistry. The detailed information about these benchmarks is as follows:

\begin{itemize}

    \item AIME2024: A benchmark derived from the 2024 American Invitational Mathematics Examination (AIME), a challenging mathematics competition aimed at high school students in the U.S., designed specifically to evaluate the advanced mathematical reasoning abilities of AI models. It consists of complex problems covering algebra, geometry, combinatorics, and number theory, each requiring integer solutions ranging from 0 to 999. Models are tested on their ability to perform multi-step reasoning, provide accurate step-by-step explanations, and derive correct final answers.

    \item AIME2025: Like AIME2024, the AIME2025 benchmark is based on the 2025 American Invitational Mathematics Examination (AIME), an advanced and highly respected mathematics competition aimed at high school students in the United States, intended specifically for evaluating the mathematical reasoning and problem-solving capabilities of AI models.

    \item AMC2023: A benchmark derived from the 2023 American Mathematics Competitions (AMC), specifically designed to evaluate the mathematical reasoning abilities of AI models. It consists of 40 questions, covering various mathematical topics such as algebra, geometry, number theory, and combinatorics.

    \item GPQA-Diamond \cite{rein2024gpqa}: A subset of the GPQA dataset, specifically designed to assess the reasoning capabilities of advanced AI systems and highly knowledgeable humans on extremely difficult, domain-expert-level questions in biology, physics, and chemistry. The "Diamond" subset is the hardest subset of the benchmark, which is intended to facilitate research on reasoning models.
\end{itemize}
We evaluated and measured these models on the above benchmarks:

\begin{itemize}
    \item QwQ-32B \cite{qwq32b}: A large-scale language model designed to achieve robust performance across a wide range of natural language processing tasks. Developed with 32 billion parameters, QwQ32B leverages advanced architecture and training techniques to enhance understanding, generation, and reasoning in general and specialized domains.

    \item Phi4-Reasoning-Plus \cite{abdin2025phi4reasoning}: Built on Phi-4 Base, it is an advanced language model specifically designed to excel in complex reasoning and problem-solving tasks across multiple domains, demonstrating strong performance in textual data.

    \item Qwen3-32B\cite{qwen3}: A state-of-the-art large language model developed by Alibaba Cloud, featuring 32 billion parameters and designed to deliver high performance across a broad spectrum of language understanding, text generation and reasoning tasks. 

    \item QwQ-32B-Preview \cite{qwq32b}: An experimental large language model developed by Alibaba, designed to advance AI reasoning capabilities. With 32.5 billion parameters and a 32,768-token context window, it is specifically tested on benchmark AIME2024 and AMC2023 to compare with other methods.
\end{itemize}

\subsection{Visual QA}
 Additionally, we incorporate evaluations on the multimodal benchmarks including \textbf{MMMU-Pro} \cite{yue2024mmmu-pro}, \textbf{MMMU} \cite{yue2023mmmu}, \textbf{Math-Vista} \cite{lu2024mathvista} and \textbf{EMMA-mini} \cite{hao2025emma-mini} to further explore the models’ capabilities across diverse reasoning and multimodal tasks.
 Here is the detailed information:
\begin{itemize}

     \item MMMU \cite{yue2023mmmu}: A multimodal evaluation benchmark specifically designed to test the capabilities of AI models on college-level tasks that require both advanced subject knowledge and deliberate reasoning across a broad range of academic disciplines. 
    \item MMMU-Pro \cite{yue2024mmmu-pro}: MMMU-Pro is an enhanced evaluation benchmark designed to rigorously test the true understanding and reasoning capabilities of multimodal AI models. Building on the original MMMU benchmark, it forces models to simultaneously process and integrate visual and textual information, simulating real-world scenarios that require human-like cognitive skills.

    \item Math-Vista \cite{lu2024mathvista} : A comprehensive benchmark specifically designed to evaluate and challenge the mathematical reasoning abilities of large language and multimodal models within visual contexts. It requires models to perform deep, fine-grained visual understanding and complex compositional reasoning across diverse mathematical tasks.

    \item EMMA-mini \cite{hao2025emma-mini}: A specialized benchmark designed to rigorously assess the ability of Multimodal Large Language Models (MLLMs) to perform integrated, organic reasoning over both text and images—an essential aspect of human intelligence. Unlike existing benchmarks that often focus on text-based reasoning or superficial visual cues, EMMA-mini presents tasks spanning mathematics, physics, chemistry, and coding, all of which require genuine cross-modal reasoning that cannot be solved by independently analyzing text or images alone.

\end{itemize}
We evaluated and measured these models on the above benchmarks:

\begin{itemize}
    \item Kimi-VL-A3B-Thinking \cite{kimiteam2025kimi-vl}: An efficient open-source vision-language model (VLM) built on a Mixture-of-Experts (MoE) architecture, designed to deliver advanced multimodal reasoning, robust long-context understanding, maths problem solving as well as strong agent capabilities.

    \item QvQ-72B-Preview \cite{qvq-72b-preview}: QVQ-72B-preview is an open-source, large-scale multimodal reasoning model built on top of Qwen2-VL-72B, achieving remarkable performance on challenging benchmarks. In this part of the experiment, the image recognition and reasoning capability of this model has been tested.
\end{itemize}
\subsection{Video QA}

Furthermore, we conduct experiment on video benchmarks, whose name and details is listed as follows:

\begin{itemize}
    \item MMVU \cite{zhao2025mmvu}:  A comprehensive dataset designed to evaluate the capabilities of AI models in understanding and reasoning over expert-level, domain-specific videos. Each example is meticulously crafted using a textbook-guided annotation process, ensuring that questions require both visual comprehension and the application of domain-specific knowledge. What's unique to MMVU is the inclusion of expert-annotated reasoning rationales and relevant domain knowledge for each question, which largely facilitates the fine-grained analysis of model performance.

    \item VSI-Bench \cite{yang2024vsi-bench}: A pioneering dataset, designed to evaluate the visual-spatial reasoning capabilities of multimodal large language models (MLLMs). It comprises many question-answer pairs derived from egocentric videos that are sourced from public indoor 3D scene reconstruction datasets, aiming to provide a comprehensive benchmark for testing and improving the spatial reasoning abilities of multimodel large language models, moving beyond traditional static image evaluations. 
\end{itemize}

\clearpage

\begin{table*}[!t]
\setlength{\tabcolsep}{8pt}
\begin{minipage}[!t]{\textwidth}
    \textbf{E \ \ \ \ \large Additional Experiment Results \& Case Study}\\[0.5em]
    
    \caption{\textbf{Complete Experiment results of \method on Qwen3 Series Models and Other Distill Models.} 
    }
    \centering
    \begin{tabular}{cllllllll}
        \toprule
         \multirow{2}{*}{\textbf{Strategy}} & \multicolumn{2}{c}{\textbf{AMC 2023}} & \multicolumn{2}{c}{\textbf{AIME 2024}} & \multicolumn{2}{c}{\textbf{AIME 2025}} & \multicolumn{2}{c}{\textbf{GPQA-D}} \\
         & ACC$\uparrow$ & LEN$\downarrow$ & ACC$\uparrow$ & LEN$\downarrow$ & ACC$\uparrow$ & LEN$\downarrow$ & ACC$\uparrow$ & LEN$\downarrow$ \\
         \midrule
         \midrule
         \multicolumn{9}{c}{\textbf{Qwen3-32B}} \\
         \midrule
         \textbf{Original} & 97.50 & 6424 & 81.33 & 12720 & 66.67 & 14987 & 69.19 & 5613\\
         \textbf{NoThink} & 59.50 & 1240 & 25.33 & 2511 & 20.00 & 2165 & 50.50 & 605\\
         \textbf{\method} & 
         96.67   & 
         5560    & 
         83.33   & 
         10732   &
         64.44   &
         12930   & 
         63.13   & 
         4788    \\
         \midrule
         
         \multicolumn{9}{c}{\textbf{Qwen3-14B}} \\
         \midrule
         \textbf{Original} & 96.25 & 6677 & 78.67 & 14217 & 78.00 & 14765 & 59.59 & 4633  \\
         \textbf{NoThink} & 69.50 & 1749 & 33.33 & 3559 & 26.67 & 3171 & 38.59 & 1001 \\
         \textbf{\method} &
         95.50   & 
         4714    & 
         73.33   & 
         10919   & 
         61.33   & 
         12104   & 
         54.75   & 
         3889    \\

         \midrule
         \multicolumn{9}{c}{\textbf{Qwen3-8B}} \\
         \midrule
         \textbf{Original} & 97.50 & 8513 & 77.33 & 14142 & 74.61 & 16094 & 57.07 & 5904  \\
         \textbf{NoThink} & 66.50 & 1760 & 28.89 & 3362 & 25.56 & 3719 & 32.93 & 1271 \\
         \textbf{\method} &
         94.50     & 
         5251      & 
         72.67     & 
         10963     & 
         60.00     &
         13674     & 
         51.71     & 
         4735      \\

         \midrule
         \multicolumn{9}{c}{\textbf{Qwen3-4B}} \\
         \midrule
         \textbf{Original} & 93.75 & 8125 & 70.00 & 13488 & 70.00 & 18086 & 53.54 & 5965  \\
         \textbf{NoThink} & 70.00 & 2236 & 33.33 & 4068 & 23.00 & 4656 & 27.27 & 1288 \\
         \textbf{\method} &
         95.50     & 
         4523      & 
         65.33     & 
         10358     & 
         56.67     & 
         12213     & 
         76.26     & 
         3178      \\

        \midrule
         \multicolumn{9}{c}{\textbf{Llama-Nemontron-Nano-8B-v1}} \\
         \midrule
         \textbf{Original} & 71.50 & 4535 & 39.33 & 7371 & 44.67 & 11798 & 54.1 & 5071 \\
         \textbf{NoThink} & 39.00 & 2982 & 6.6 & 2185 & 14.67 & 3677 & 30.30 & 3109 \\
         \textbf{\method} &
         72.00  & 
         3690   & 
         30.67  & 
         4865   & 
         33.33  & 
         7271   & 
         42.83  & 
         3754    \\
         
         \midrule
         \multicolumn{9}{c}{\textbf{Deepseek-R1-Distill-Qwen-7B}} \\
         \midrule
         \textbf{Original} & 73.00 & 4796 & 34.67 & 7755 & 40.00 & 13767 & 49.10 & 3809 \\
         \textbf{NoThink} & 30.00 & 2552 & 18.67 & 3895 & 10.00 & 2724 & 21.00 & 1112 \\
         \textbf{\method} &
         72.00    & 
         4315     & 
         26.67    & 
         7247     & 
         31.33    & 
         8236     & 
         40.91    & 
         3672      \\

         \bottomrule
    \end{tabular}
\label{tab_sft_result}
\end{minipage}
\end{table*}

\begin{figure*} 
    \centering
    \begin{subfigure}[b]{0.48\textwidth}
        \includegraphics[width=\linewidth]{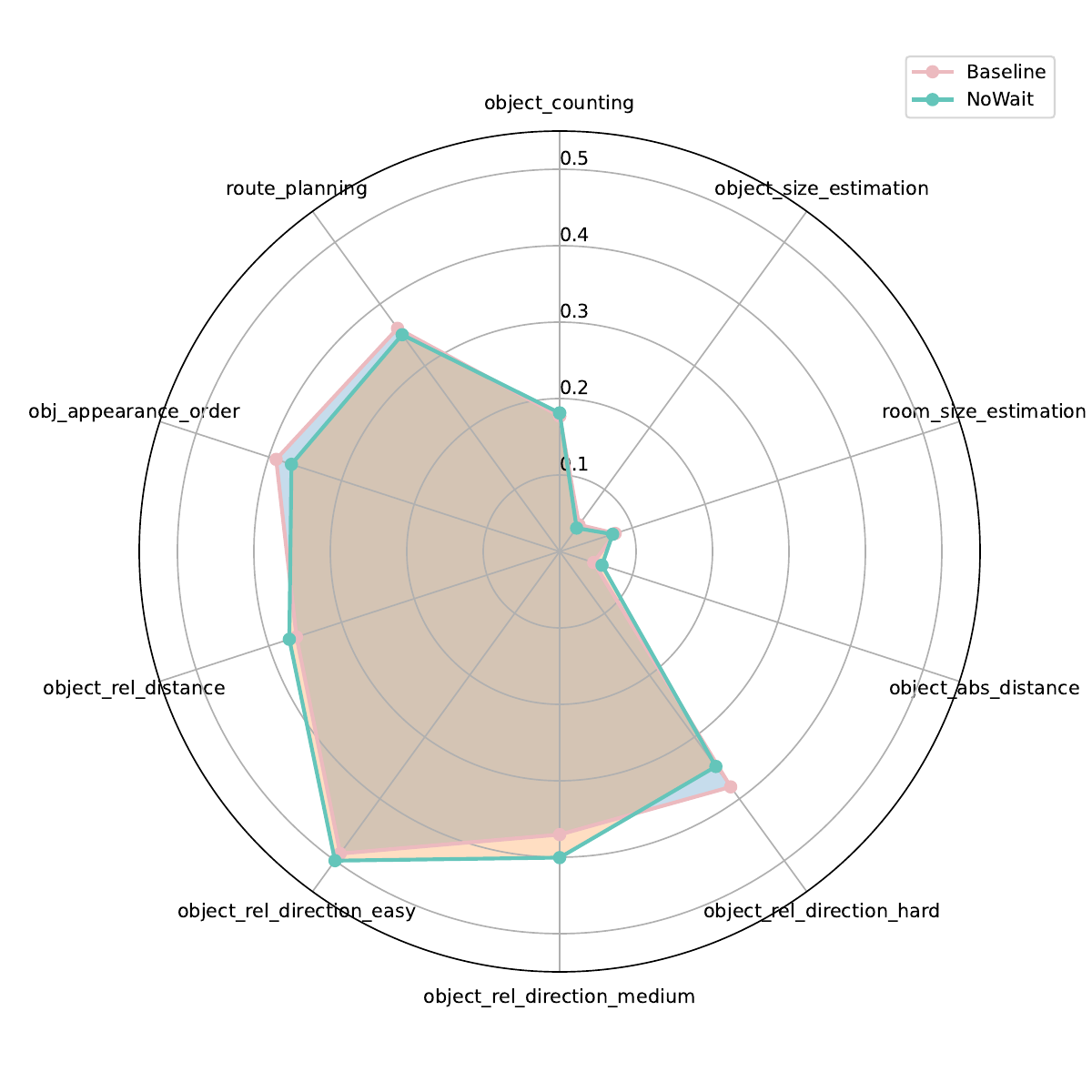}
        \caption{\textbf{Accuracy Radar Map for QvQ-72B-Preview on VSI-Bench}}
    \end{subfigure}
    \hfill
    \begin{subfigure}[b]{0.48\textwidth}
        \includegraphics[width=\linewidth]{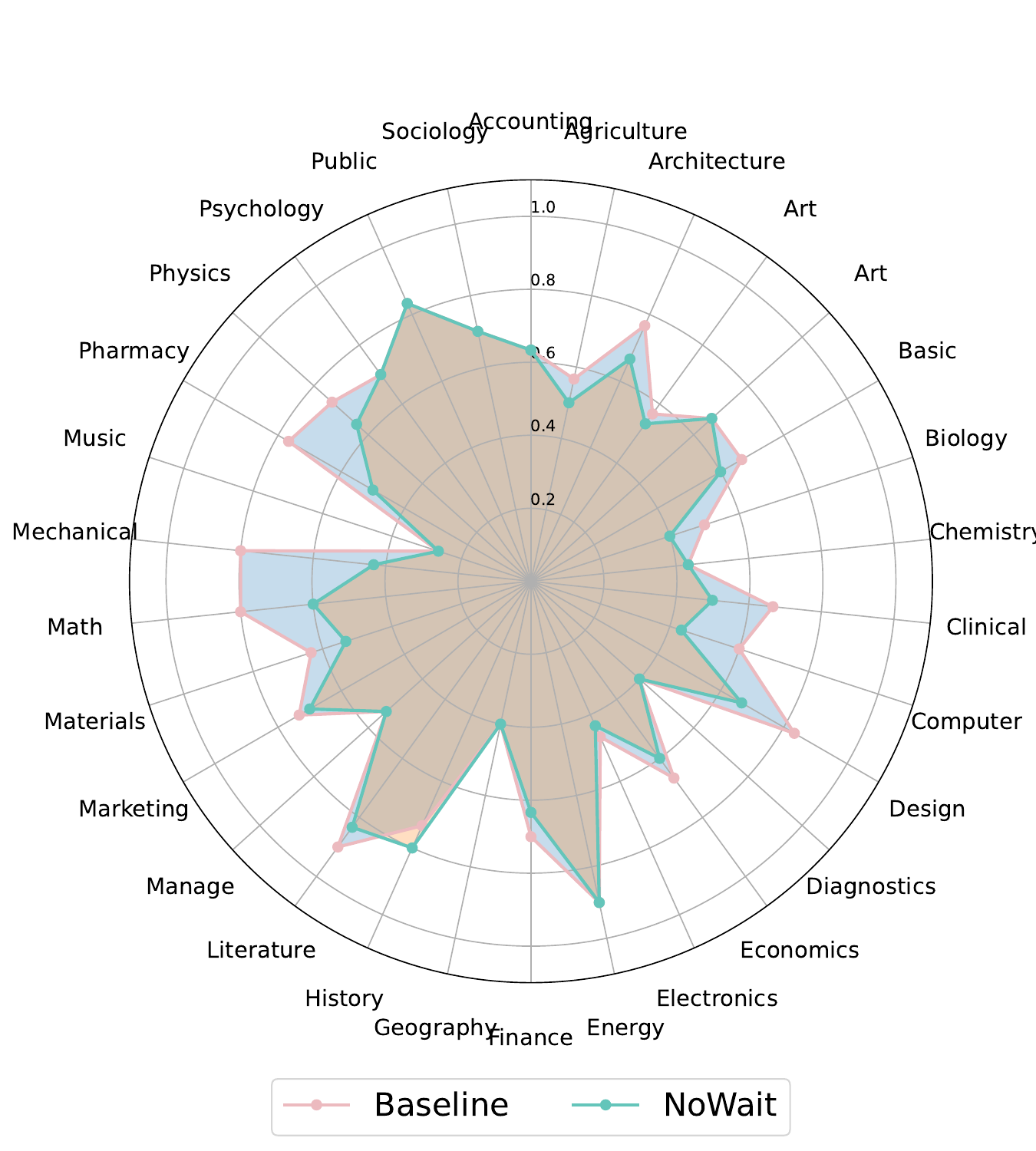}
        \caption{\textbf{Accuracy Radar Map for Kimi-VL-A3B-Thinking-7B on MMMU}}
    \end{subfigure}
\end{figure*}

\begin{figure*}[!t]
    \centering
    \includegraphics[width=\textwidth]{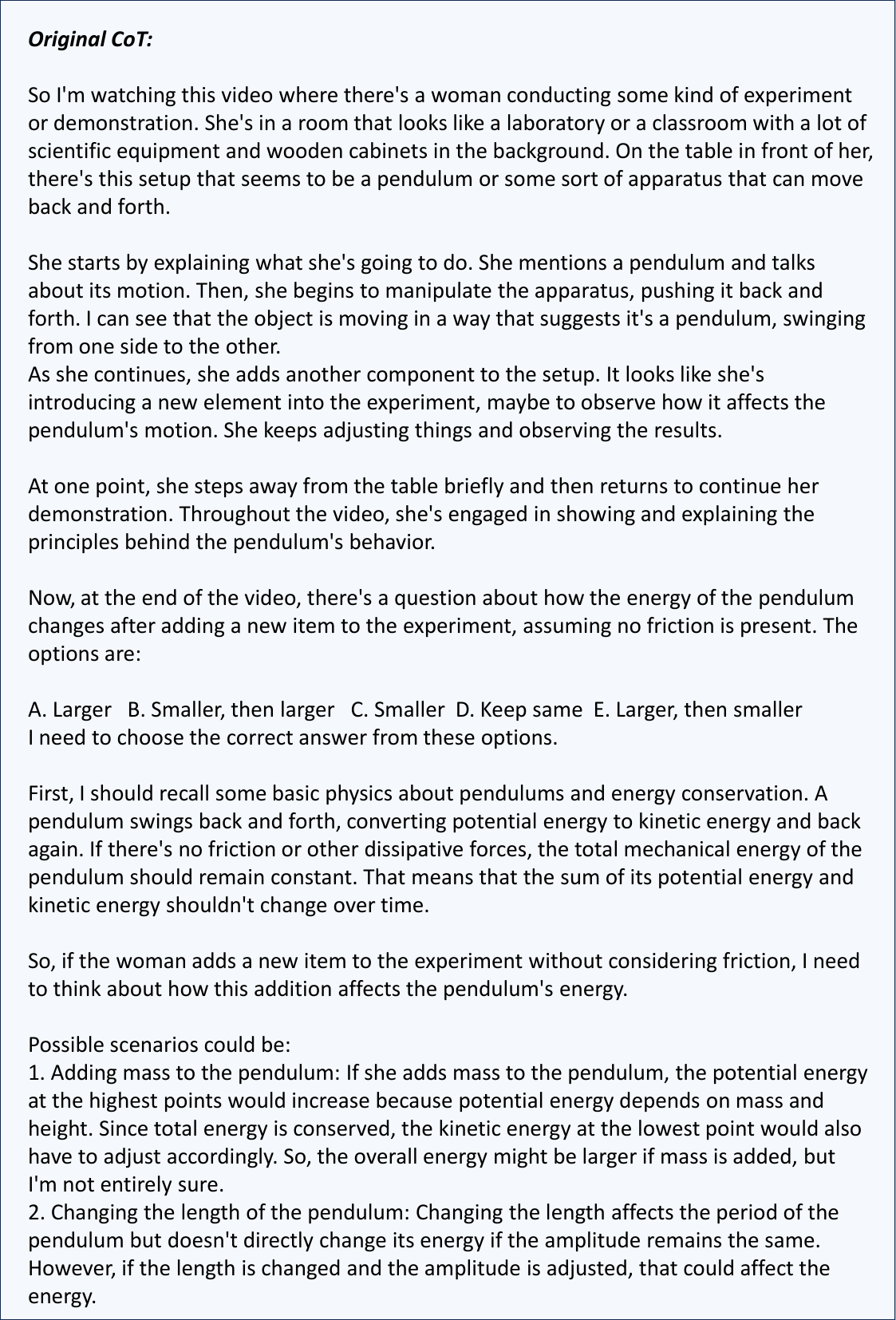}

    \caption{\textbf{A CoT Example from QvQ-72B-Preview on MMVU 2023.}}
    \label{fig:video_Baseline_Cot}
\end{figure*}

\begin{figure*}[!t]
    \centering
    \includegraphics[width=\textwidth]{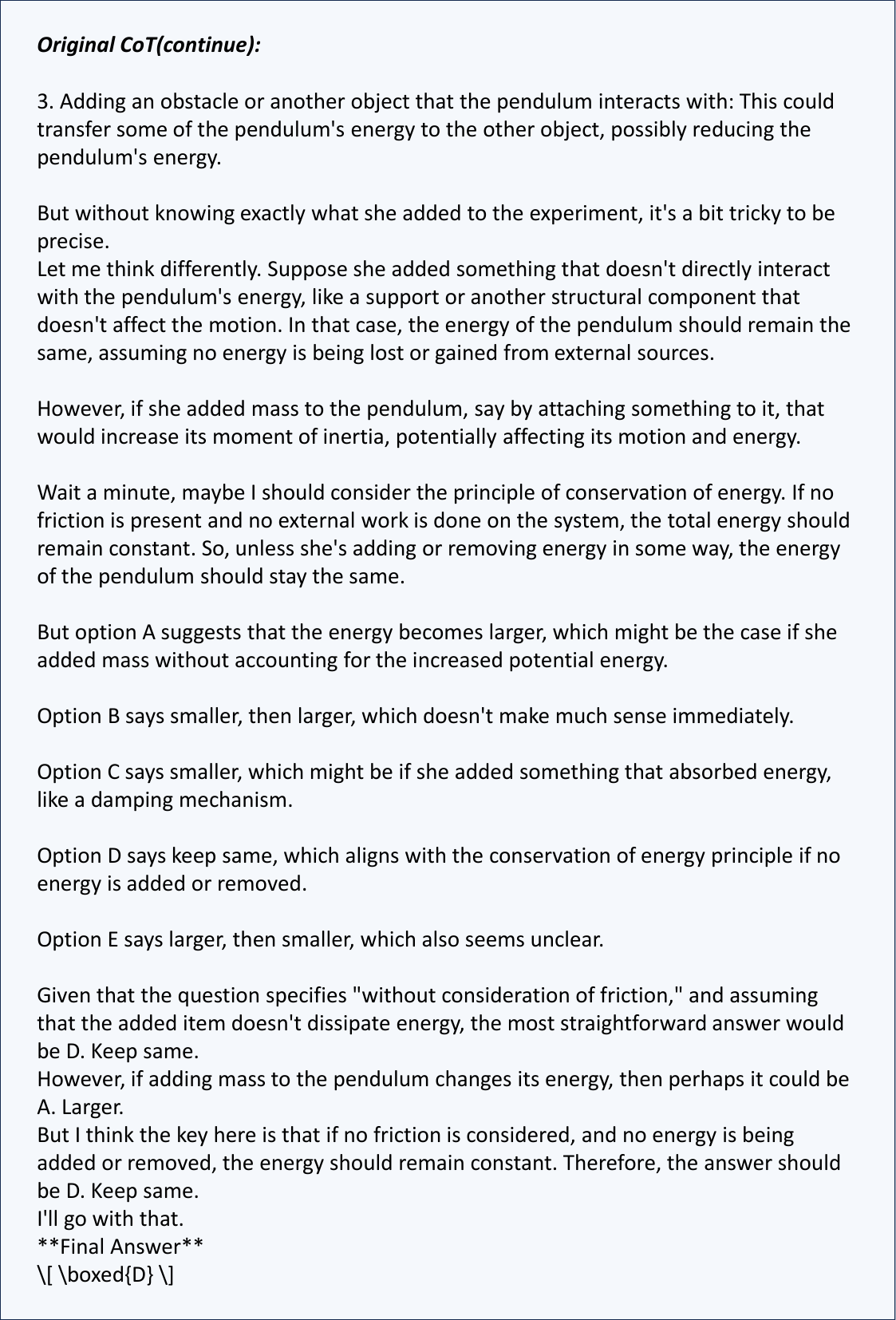}

    \caption{\textbf{A CoT Example from QvQ-72B-Preview on MMVU 2023.}}
    \label{fig:video_Baseline_Cot1}
\end{figure*}

\begin{figure*}[!t]
    \centering
    \includegraphics[width=\textwidth]{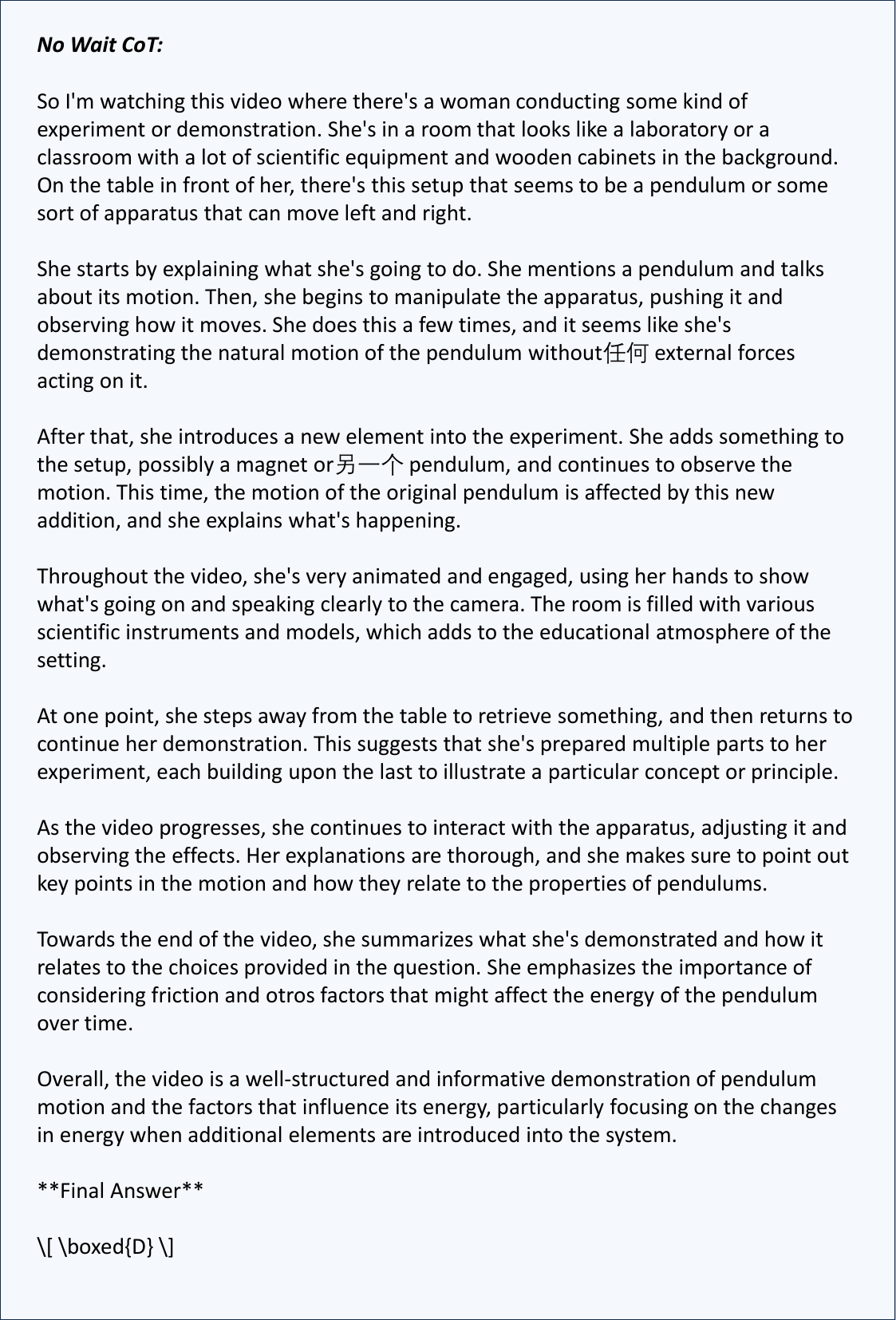}
    \caption{\textbf{A CoT Example from QvQ-72B-Preview applied \method on MMVU 2023.}}
    \label{fig:video_NoWait_Cot}
\end{figure*}

\begin{figure*}[!t]
    \centering
    \includegraphics[width=\textwidth]{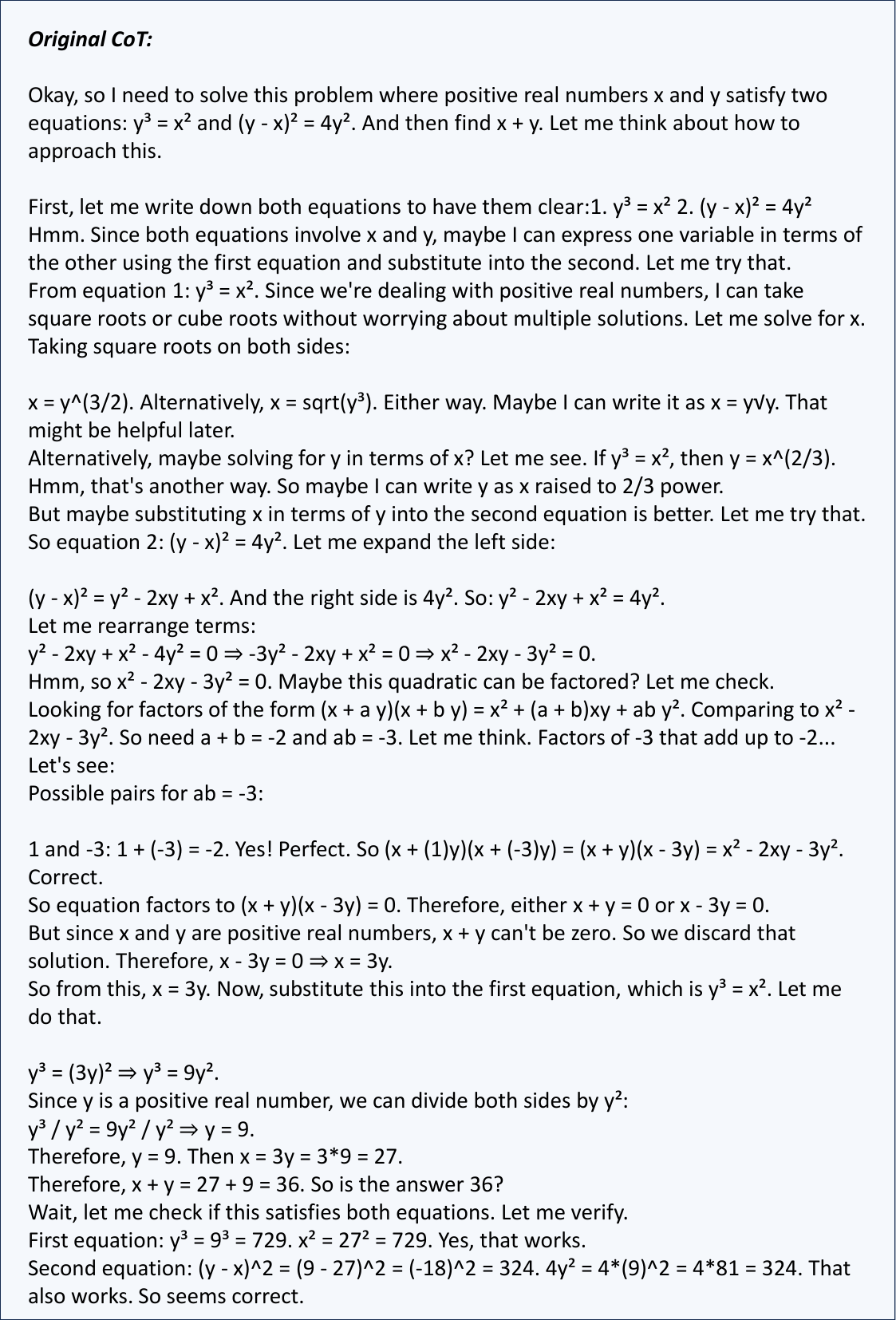}
    \caption{\textbf{A CoT Example from Qwen3-32B on AMC 2023.}}
    \label{fig:AMC23-original1}
\end{figure*}

\begin{figure*}[!t]
    \centering
    \includegraphics[width=\textwidth]{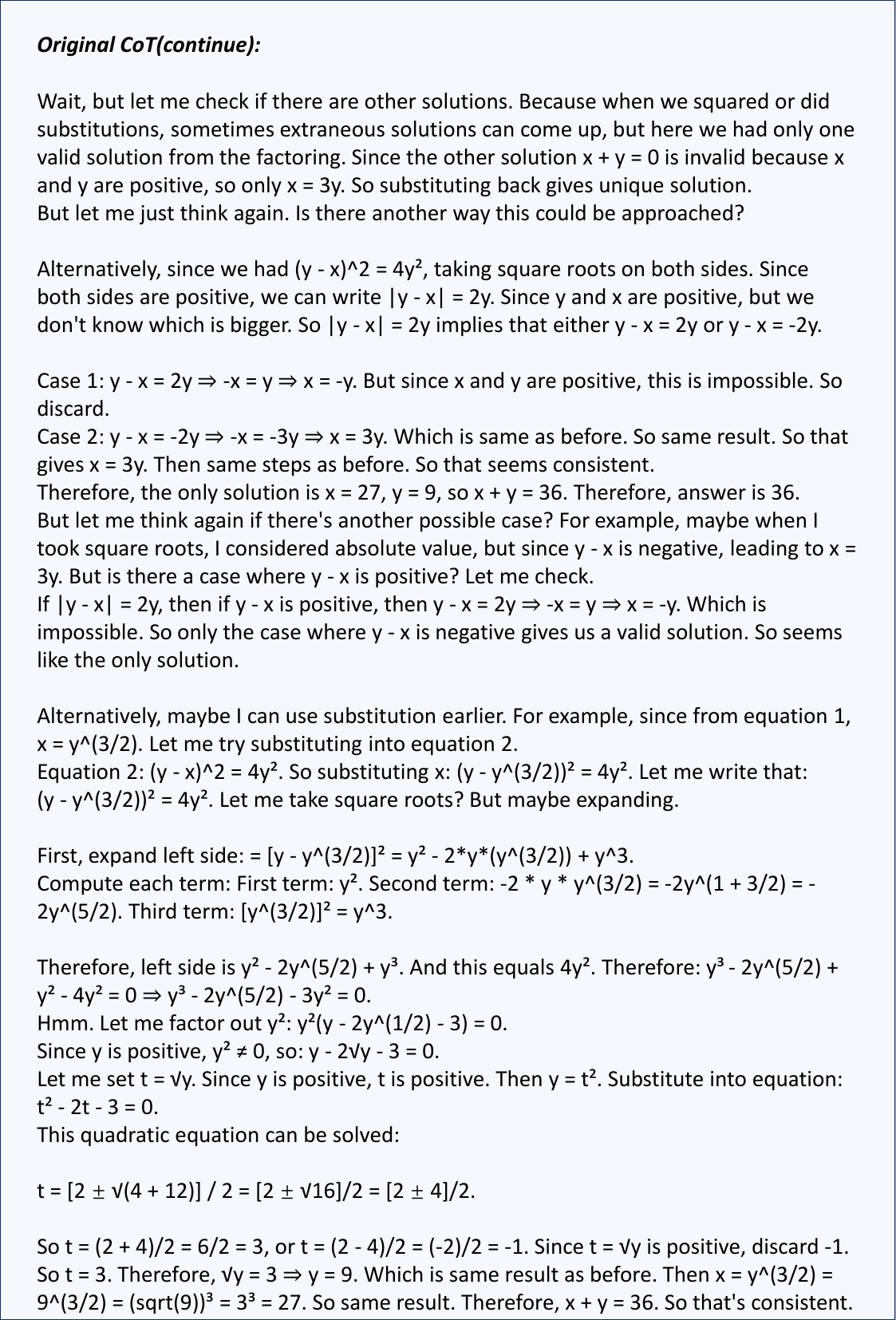}
    \caption{\textbf{A CoT Example from Qwen3-32B on AMC 2023.}}
    \label{fig:AMC23-original2}
\end{figure*}

\begin{figure*}[!t]
    \centering
    \includegraphics[width=\textwidth]{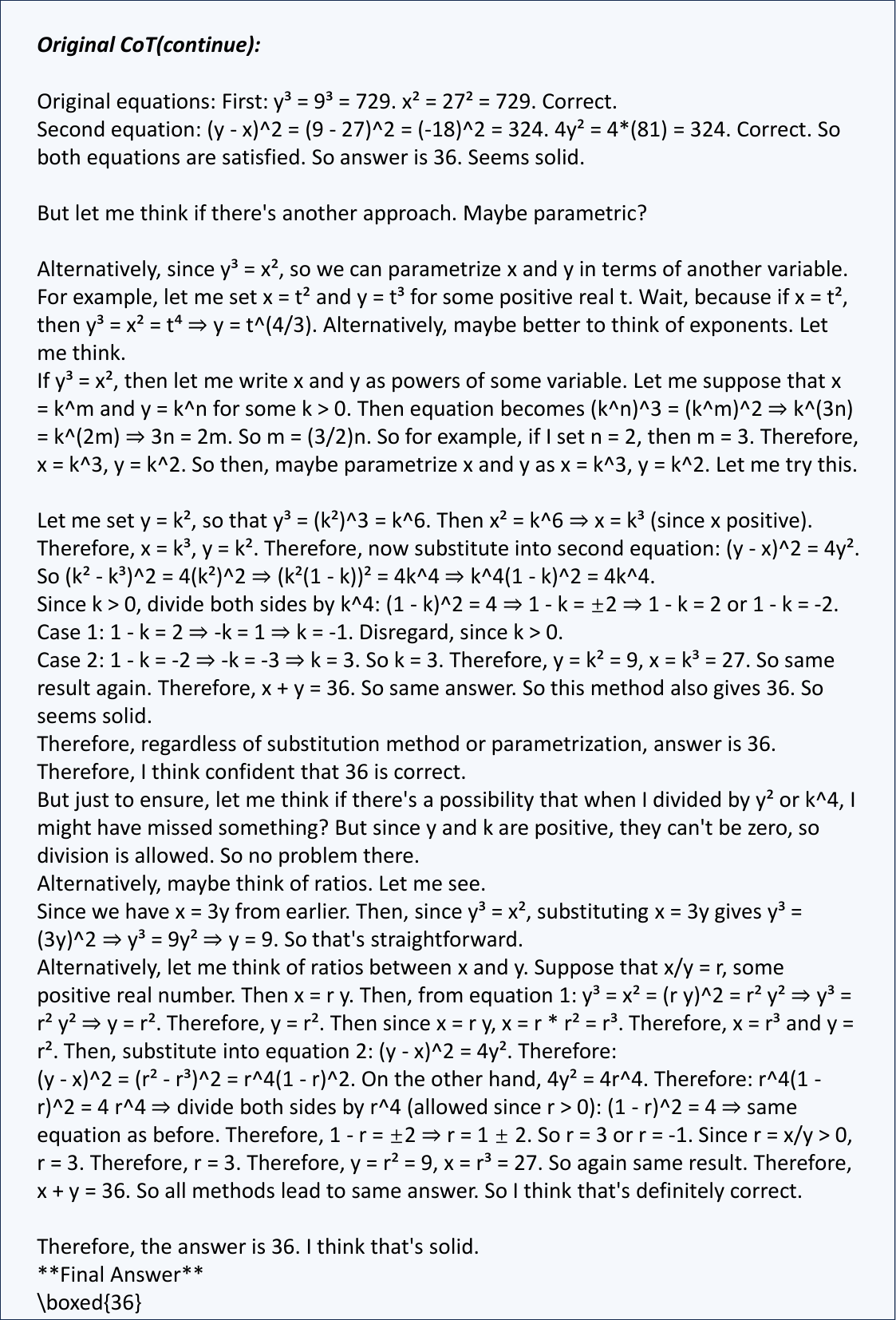}
    \caption{\textbf{A CoT Example from Qwen3-32B on AMC 2023.}}
    \label{fig:AMC23-original3}
\end{figure*}

\begin{figure*}[!t]
    \centering
    \includegraphics[width=\textwidth]{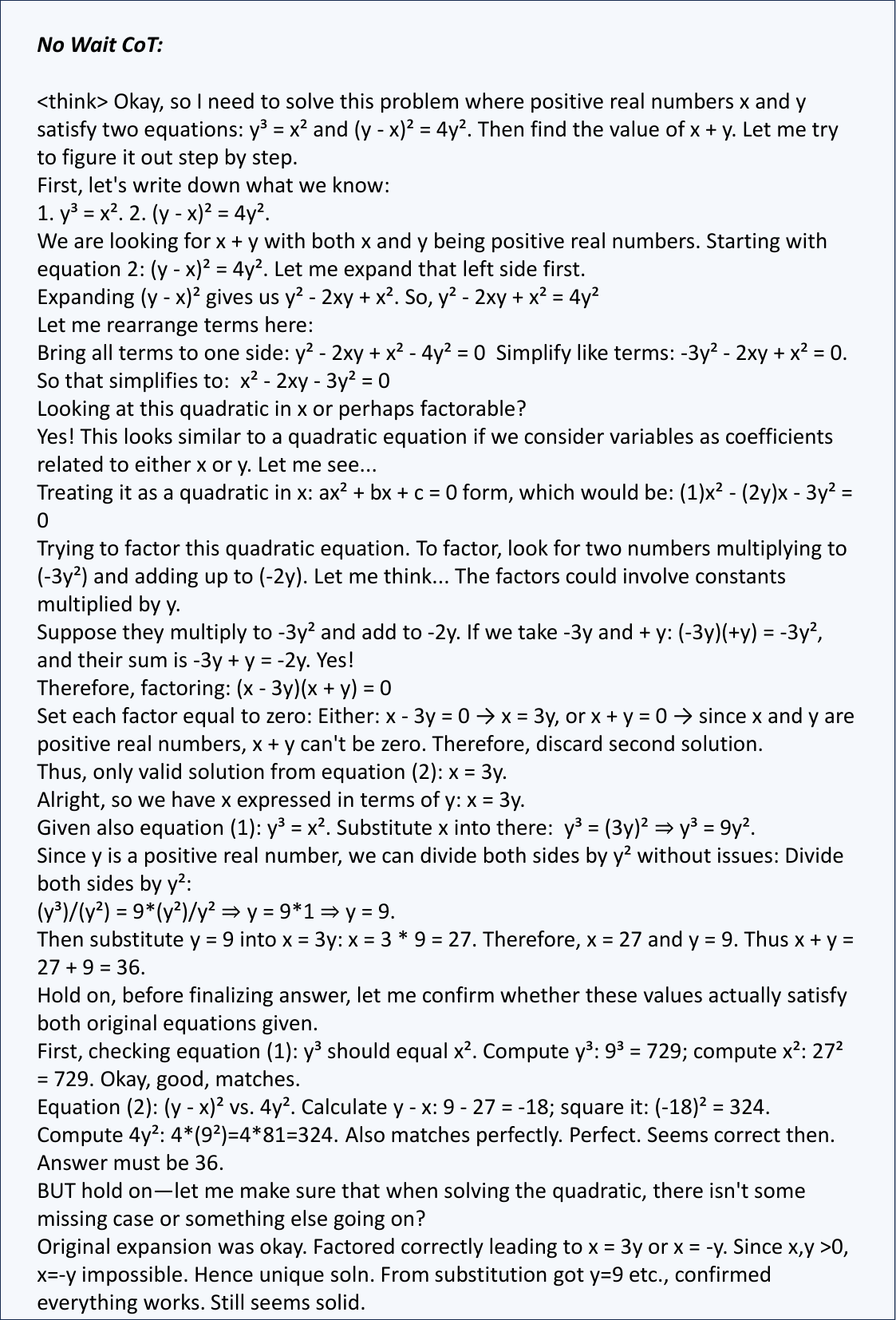}
    \caption{\textbf{A CoT Example from Qwen3-32B applied \method on AMC 2023.}}
    \label{fig:AMC23-NOWAIT1}
\end{figure*}

\begin{figure*}[!t]
    \centering
    \includegraphics[width=\textwidth]{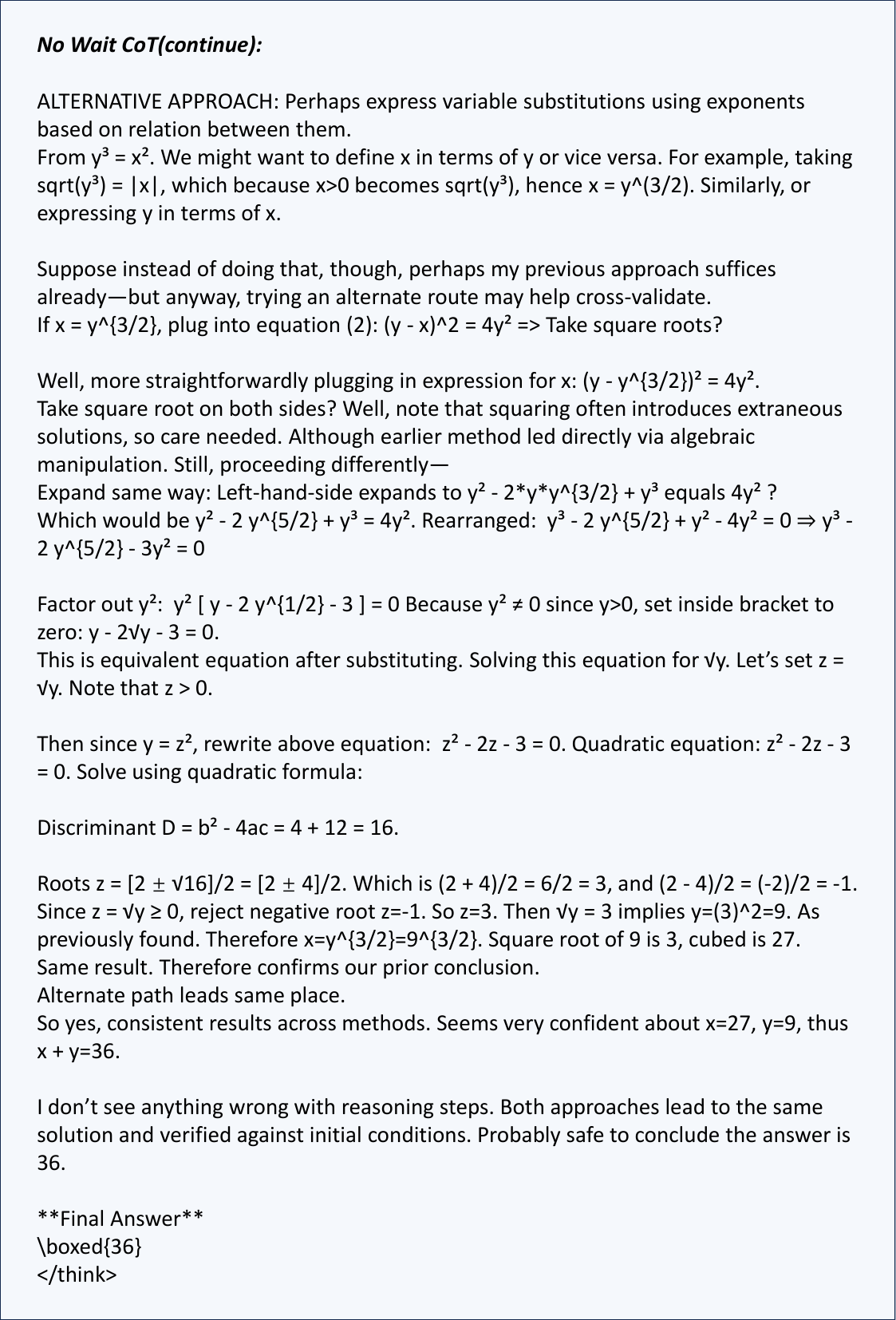}
    \caption{\textbf{A CoT Example from Qwen3-32B applied \method on AMC 2023.}}
    \label{fig:AMC23-NOWAIT2}
\end{figure*}

\end{document}